\documentclass{article}

\usepackage{arxiv}

\usepackage[utf8]{inputenc} 
\usepackage[T1]{fontenc}    
\usepackage{url}            
\usepackage{booktabs}       
\usepackage{amsfonts}       
\usepackage{nicefrac}       
\usepackage{microtype}      
\usepackage{lipsum}		
\usepackage{graphicx}
\usepackage[square,numbers]{natbib}
\usepackage{doi}
\usepackage{amsmath}

\usepackage{algorithm, algpseudocode}
\usepackage{svg}
\usepackage{hyperref}

\newtheorem{remark}{Remark}

\title{Busemann energy-based attention for emotion analysis in Poincar\' e discs}

\author{Zinaid Kapi\'c \\
	Faculty of Engineering \\
	University of Rijeka \\ 
	Vukovarska 58, 51000 Rijeka \\
	Croatia\\
	\texttt{zkapic@uniri.hr}
	\And
	Vladimir Ja\'cimovi\'c \\
    Faculty of Natural Sciences and Mathematics \\ 
    University of Montenegro \\
    Cetinjski put, bb., 81000 Podgorica \\
    Montenegro \\
	\texttt{vladimirj@ucg.ac.me} \\
}

\renewcommand{\shorttitle}{Busemann energy-based attention for emotion analysis in Poincar\' e discs}

\hypersetup{
pdftitle={Busemann energy-based attention for emotion analysis in Poincar\' e discs},
pdfsubject={Computer Science},
pdfauthor={Zinaid Kapi\'c, Vladimir Ja\'cimovi\'c},
pdfkeywords={Hyperbolic representations, Affective computing, M\" obius transformation, Circumplex model, Semantic ambiguity},
}

\begin{document}
\maketitle

\begin{abstract}
	We present EmBolic - a novel fully hyperbolic deep learning architecture for fine-grained emotion analysis from textual messages. The underlying idea is that hyperbolic geometry efficiently captures hierarchies between both words and emotions. In our context, these hierarchical relationships arise from semantic ambiguities. EmBolic aims to infer the curvature on the continuous space of emotions, rather than treating them as a categorical set without any metric structure. In the heart of our architecture is the attention mechanism in the hyperbolic disc. The model is trained to generate queries (points in the hyperbolic disc) from textual messages, while keys (points at the boundary) emerge automatically from the generated queries. Predictions are based on the Busemann energy between queries and keys, evaluating how well a certain textual message aligns with the class directions representing emotions. Our experiments demonstrate strong generalization properties and reasonably good prediction accuracy even for small dimensions of the representation space. Overall, this study supports our claim that affective computing is one of the application domains where hyperbolic representations are particularly advantageous.
\end{abstract}


\section{Introduction}
Hyperbolic data representations emerged as a powerful paradigm in machine learning (ML) about a decade ago. Initial interest in hyperbolic ML was to a certain extent inspired by studies on low-distortion embedding of the trees in the hyperbolic plane \cite{sarkar2011low} and on the negative curvature of complex networks \cite{krioukov2010hyperbolic}. The underlying idea is that hyperbolic geometry naturally captures important structural information of some ubiquitous datasets. Hierarchies, hypernyms and contextual uncertainties are better encoded in the negative curvature manifolds, thus enabling more compact and efficient ML models. An example of the data where hierarchical relations play an essential role are natural languages. Having this in mind, it is not surprising that one of the pioneering case studies in hyperbolic ML dealt with the word representations \cite{nickel2017poincare}. Authors reported a drastic dimensionality reduction for word embeddings in hyperbolic balls compared to Euclidean spaces. This study motivated extensive research efforts on hyperbolic models in natural languages \cite{tifrea2018poincar,leimeister1809skip,nickel2018learning,sala2018representation} that recently advanced towards hyperbolic LLM's \cite{patil2025hyperbolic}.

In parallel, affective computing appeared as a branch of computer science in 1990's \cite{picard2000affective}. With the advent of AI this field provided a conceptual framework for impressive advances in Emotion AI. 

The present study argues that Emotion AI is one of the fields where hyperbolic ML can unleash its great potential and prove its effectiveness compared to the standard Euclidean models. We present EmBolic - a fully hyperbolic deep learning architecture trained to recognize emotions from textual messages. Fine-grained emotion analysis is known to be a challenging problem in ML, due to subtle semantic interplay between words and emotions. EmBolic is inspired by classical models of affective states and leverages rigorous geometric techniques in order to recognize emotional context from the text. In whole, EmBolic maps both words and emotions to hyperbolic manifolds and its training boils down to learning mappings between these manifolds.

\subsection{Modeling assumptions}

\label{subsection_modeling_assumption}
Existing emotion analysis models consider a finite set of emotions without assuming any metric structure. Such models typically output a categorical probability distribution over a finite set of outcomes. The probabilities are generated from the latent representations using the softmax function. Such an approach is based on an apparent simplification, as it does not take into account relationships between emotions.  

A more sophisticated approach would consider a continuous space of emotions. Evidences suggest that, similar to the color spectrum, emotions lack exact borders that clearly differentiate between them \cite{russell1994fuzzy}. This observation raises the key question regarding the geometry of emotions, that is - on the metric tensor and curvature. Addressing this question, we recall the classical circumplex model of affect \cite{russell1980circumplex}, proposed by James Russell. Emphasize the Russell's map of emotions is not imposed in the architecture. Instead, circumplex models serve solely as an inspiration and a kind of theoretical justification for a priori assumptions regarding geometry of the latent space. Instead, the model is trained to learn the emotion map arising from the annotated data.

In whole, two basic modeling assumptions constituting the conceptual framework for our model can be substantiated as follows:

\begin{description}
	\item[A1)] Words are represented by points in the unit hyperbolic (multi-)disc.
	\item[A2)] Emotions are represented by points in the unit hyperbolic (multi-)disc.
\end{description}

As explained above, the first assumption is widely adopted in the ML community and does not require an additional justification. The motivation for the second assumption stems from a long track of psychological research on the circumplex model of affect. This model and its extensions will be clarified in the next subsection.

As hyperbolic discs and multi-discs serve as latent spaces for our architecture, they will be briefly explained in the next Section. 

In summary, EmBolic does not impose any a priori assumptions on distances between emotions. In such a way, we avoid bias as much as possible, so that the model can reexamine and refine assumption A2.

\subsection{Circumplex model and its applications in affective computing}

In his seminal study Russell \cite{russell1980circumplex}, asked participants to sort 28 emotion words into categories based on perceived similarity. He further grouped the ratings based on correlations and obtained the mapping of eight basic emotions uniformly along the circle, with the angle of 45 degrees between each pair of neighboring emotions. This map immediately unveils two perpendicular axes interpreted as arousal and valence content of each emotion. This implies that antipodal (polar) emotions have opposite values of both arousal and valence. These findings have been substantiated into the famous Russell's circumplex model. The key hypothesis arising from this model is that affective states arise from cognitive interpretations of core neural sensations that are the product of two independent neurophysiological systems. The original Russell's paper treated relationship between words and emotions in English language. The follow-up study \cite{russell1989cross} investigated four languages and found empirical evidence that the circular mapping remains valid for non-English cultural environments as well.

\begin{figure}[t]
	\centering
	\includegraphics[width=0.5\textwidth]{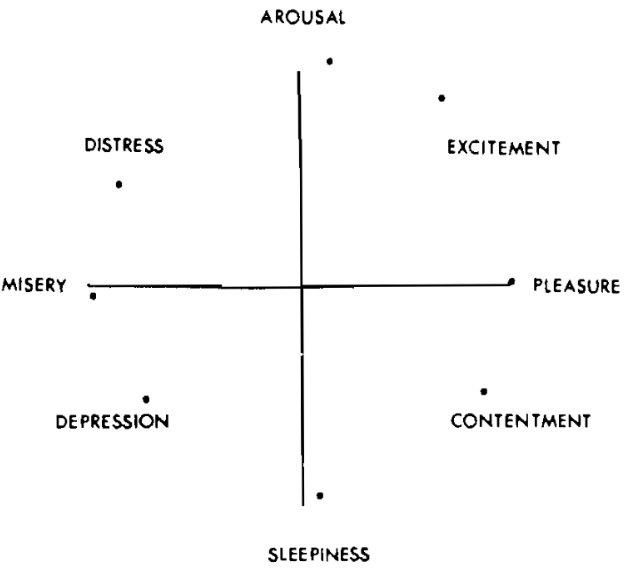}
	
	\caption{Circumplex model of affect by \citet{russell1980circumplex}. Permissions obtained.}
	\label{fig1}
\end{figure}

From the mathematical point of view, Russell's model suggests that the space of emotions seems to exhibit a positive curvature (circle, torus or a sphere). Further investigations repeatedly yielded two-dimensional models of affective experience, with the two coordinates conceptualized in different ways (arousal-valence, approach-withdrawal, tension-energy). Extensions of the Russell's model placed affective states in the interior of the circle, thus encoding hierarchical relationships between emotions. For instance, Klaus Scherer introduced a semantic space where emotions are represented by points in the disc \cite{scherer2005emotions}. This disc of emotions is usually referred to as Scherer's circumplex model. Figure \ref{fig1} shows Russell's circumplex model of affective states as presented in his paper.

With the explosive growth of supercomputing resources and ML techniques circumplex models proved their relevance as a framework for computational tools and Emotion AI technologies.

For instance, the paper \cite{paltoglou2012seeing} introduced classification methods based on support vector machines for predicting levels of arousal and valence trained on the dairy-like blog posts.

Furthermore, Scherer with collaborators \cite{scherer2013grid} developed an instrument named Geneva Emotion Wheel for measuring emotional reactions to objects, events and situations.

Circumplex models have also been deployed for the emotion analysis from facial expressions. The paper \cite{ahn2010asymmetrical} proposed a method for computing emotional parameters from facial expressions using the circular diagram.

Another example is a facial Emotion AI technology developed by the MorphCast company \cite{TomasiCircumplex2024}.

In conclusion, the Russell's circumplex model and its extensions imply that the unit disc might be the most appropriate manifold for mapping affective states. To our best knowledge, the question of metric on this disc has never been tackled in a rigorous mathematical way. In the present study, this arises as the central modeling paradigm, addressed by the assumption A2 above. Due to the topological properties of the disc and hierarchical relationships between emotions, it is reasonable to assume that this disc is equipped with the (Poincar\' e) hyperbolic metric. Adopting this assumption, one could expect that ambivalent emotions (such as indifference) will be placed near the center of the disc, and strongly pronounced affective states close to the boundary circle.

\subsection{Related work on fine-grained emotion analysis}

Fine-grained emotion analysis is a challenging NLP task. Unlike sentiment analysis, which identifies whether a text is positive or negative, fine-grained emotion analysis detects and differentiates a large range of emotion categories that are often overlapping (e.g. joy vs excitement). The task is sensitive to contextual cues and linguistic phenomena, such as sarcasm, punctuation and informal language. Fine-grained emotion analysis is applied in various applications, among others in empathetic chatbots, human-machine interactions, and psychological assessment.

Previous research on emotion analysis has introduced datasets from diverse domains, including 
news headlines \cite{strapparava2007semeval}, tweets \cite{mohammad2018semeval}, and narratives \cite{liu2019dens}. These datasets are mostly based on coarse emotion taxonomies drawn from Ekman \cite{ekman1992argument} or Plutchik \cite{plutchik1980general}. The GoEmotions dataset \cite{demszky2020goemotions} addresses these limitations by providing 58k manually annotated Reddit comments labeled with 27 emotions plus a neutral category, making it a widely used benchmark.

Most existing approaches represent both text and emotion categories in Euclidean space. Early work relied on lexicon-based features and manually designed classifiers, often in domain-specific settings such as health-related text \cite{khanpour2018fine}. Neural models enabled data-driven learning of emotion representations. Transformer-based encoders, including BERT and RoBERTa, are now widely used for emotion analysis \cite{devlin2019bert, wang2024large}. Researchers have further improved these models with techniques such as syntactic self-attention \cite{yin2020sentibert}, emotion-aware pretraining \cite{sosea2021emlm}, and contrastive or label-aware objectives \cite{suresh2021not, yang2023cluster}. Other studies have extended Euclidean models to capture structured relationships among emotion labels. For example, \cite{yu2024emotion} propose an emotion-anchored contrastive framework for conversational data, and \cite{zhang2024message} uses graph neural networks over token representations to model semantic and temporal patterns.

Hyperbolic representations have been explored in several recent studies. HypEmo \cite{chen2023label} embeds emotion labels in hyperbolic space using a predefined hierarchy, while \cite{kumar2025semantic} propose a contrastive learning model that aligns text and label embeddings without assuming an explicit hierarchy. These approaches typically adopt a hybrid design, combining Euclidean text encoders like BERT or RoBERTa with hyperbolic label representations.

\subsection{Our contribution}

We present a fully hyperbolic emotion analysis DL model named EmBolic. The model is based on the rigorous mathematical framework including mappings and optimization methods in hyperbolic spaces. In particular, we implement a sort of attention mechanism based on the notion of (weighted) conformal barycenter in the hyperbolic disc.

Our results demonstrate that hyperbolic geometry captures hierarchical structure of both words and emotions, thus presenting a strong argument in favor of hyperbolic representations in Emotion AI.

We train our model on the fraction of the Google's GoEmotions dataset and obtain very promising results in moderate dimensions. As a collateral result, we validate our a priori assumption about hyperbolic geometry of the emotion space.

\subsection{Outline}

The next Section contains some mathematical preliminaries which are crucial for understanding of our model. We explain the Poincar\' e (multi-)disc, and the notion of the conformally invariant mean (conformal barycenter) in the hyperbolic disc. We also present the method for generation of random points in the Poincar\' e disc which is necessary for the initialization of optimization algorithms.

In Section \ref{method} we briefly explain the dataset, and continue with the EmBolic architecture, including crucial layers, scoring system and loss function.

Section \ref{results} contains results, discussion on mispredictions and the map of emotions learned during the training. Section \ref{conclusion} contains some concluding remarks and an outlook for further integration of hyperbolic DL models with Emotion AI.

The paper also contains two appendices containing more details on mathematical procedures for word embeddings, as well as a more detailed visual information on representations.

\section{Mathematical preliminaries}
\label{mathematical_preliminaries}
In standard Euclidean DL models the latent space is almost always real vector space equipped with the standard inner product. The dimension of this representation space is typically fixed. In contrast, when designing a hyperbolic DL model there are several different options for the latent space. 

The Poincar\' e disc is minimal model of hyperbolic geometry. Unit Poincar\' e disc (isomorphic to the hyperbolic half-plane) is the two-dimensional manifold with the constant curvature equal to -1. 

As two-dimensional latent space is insufficient for most real-world tasks, one can consider several higher-dimensional hyperbolic manifolds. In particular, there are two non-equivalent models of hyperbolic balls: Poincar\' e balls in real vector spaces and Bergman balls in complex vector spaces. Another popular choice in hyperbolic ML is the Lorentz group (the upper sheet of hyperboloid) equipped with the Minkowski metric, as some researchers found that this manifold is less prone to numerical instabilities during optimization.

In the present paper we embed the data (words and emotions) in hyperbolic multi-disc, i.e. in the product of Poincar\' e discs. Namely, we will represent each word as a point in the product of three Poincar\' e discs, which makes the dimension of the representation space equal to six. In the present Section we briefly expose mathematical notions and techniques that are essential for our DL model.

\subsection{Poincar\' e disc}

Consider the unit disc in the complex plane where each point is represented by a complex number $z$, such that $|z|<1$. Using complex numbers, introduce the distance between two points by the following formula 
\begin{equation}
	\label{Poincare_distance}
	d_h(z_1,z_2) = arccosh \left( \frac{|z_1-z_2|^2 + 2 (1-|z_1|^2)(1-|z_2|^2)}{2(1-|z_1|^2)(1-|z_2|^2))} \right).
\end{equation}
Metric \eqref{Poincare_distance} introduces the constant negative curvature equal to $-1$ on the unit disc. This two-dimensional manifold is named the Poincar\' e disc. We will denote it by ${\mathbb B}^2$.

\begin{figure}[htbp]
	\centering
	\includegraphics[width=0.48\columnwidth, height=0.4\textwidth, keepaspectratio]{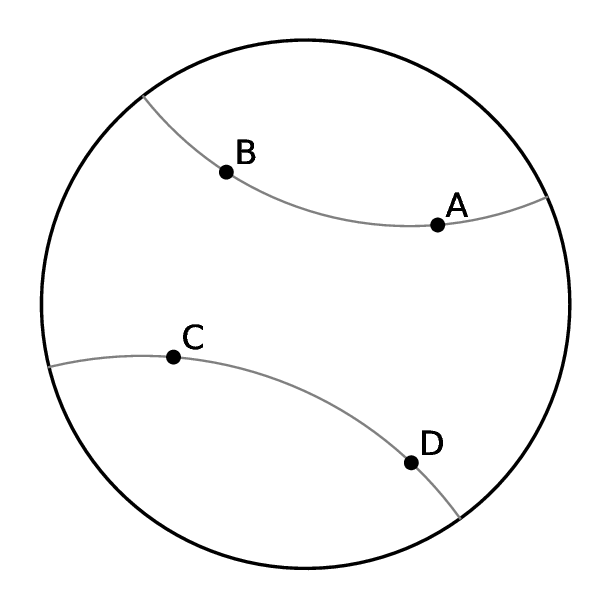}
	\hfill
	\includegraphics[width=0.48\columnwidth, height=0.4\textwidth, keepaspectratio]{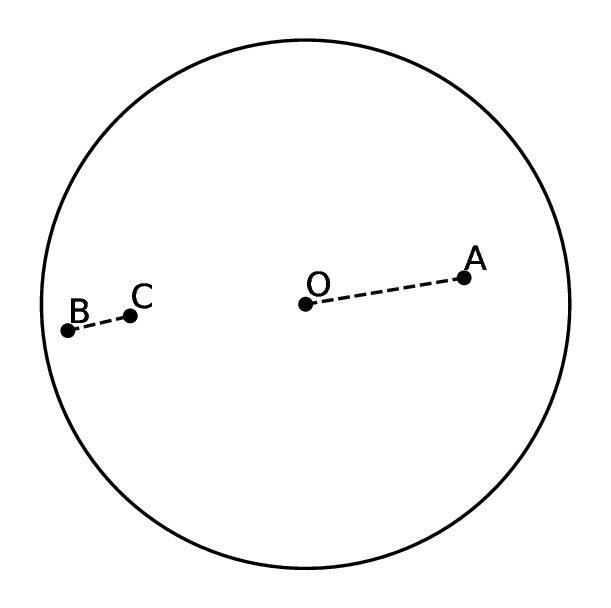}
	\caption{Visualization of the Poincar\'e disc. Two parallel geodesic lines (the shortest paths connecting points) in the left panel and visualization of distances in the right panel. The distance between points O and A is equal to the distance between B and C.}
	\label{fig2}
\end{figure}

The associated metric tensor on ${\mathbb B}^2$ is given by:
\begin{equation}
	\label{tensor}
	ds^2 = \frac{4 |dz|^2}{(1-|z|^2)^2}.
\end{equation}
Figure \ref{fig2} shows geodesics and distances in the Poincar\' e disc.

Further, consider the group of M\" obius transformations acting on the complex plane of the following form
\begin{equation}
	\label{Mobius}
	g_a(z) = e^{i \theta} \frac{a-z}{1-\bar a z}, \quad \theta \in [0,2 \pi), \; a \in \mathbb{B}^2.
\end{equation}
It is easy to verify that M\" obius transformations of the form \eqref{Mobius} map the unit disc onto itself. Hence, they form the subgroup within the group of all M\" obius transformations acting on the complex plane.

The (sub)group of M\" obius transformations is strongly related with the Lie group $SU(1,1)$ of matrices of the form:
\begin{equation*}
	\left(
	\begin{array}{cc}
		a & b \\
		- \bar b & \bar a
	\end{array}
	\right),
	\mbox{  where  } a,b \in \mathbb{C}, \quad |a|^2 + |b|^2 = 1. 
\end{equation*}
More precisely, the group of transformations of the form \eqref{Mobius} is isomorphic to the quotient group $PSU(1,1) = SU(1,1) / \pm I$. We denote this group by ${\mathbb G}$. It is well known \cite{needham2023visual} that ${\mathbb G}$ is the isometry group of the manifold ${\mathbb B}^2$. In other words, transformations \eqref{Mobius} preserve metric \eqref{Poincare_distance}.


Product of $k$ Poincar\' e discs ${\mathbb B}^2 \times \cdots \times {\mathbb B}^2$ is the $2k$-dimensional manifold named hyperbolic multi-disc. Each point in the $2k$-dimensional hyperbolic disc is represented by $k$ complex numbers $(z_1,\dots,z_k)$, where $|z_i|<1$ for $i=1,\dots,k$. The distance between two points $z=(z_1,\dots,z_k)$ and $\xi=(\xi_1,\dots,\xi_k)$ in the hyperbolic disc is
\begin{equation*}
	\rho_h(z,\xi) = \frac{1}{k} \sum_{i=1}^k d_h(z_i,\xi_i),
\end{equation*}
where $d_h(\cdot,\cdot)$ is defined in \eqref{Poincare_distance}.

\subsection{Conformal barycenter in the Poincar\' e disc}
\label{subsection_conformal_barycenter}
Computing the (weighted) average of vectors is probably the most frequently used computational procedure in ML. Unlike the Euclidean case, notion of the mean in hyperbolic geometry is not evident. In this subsection we provide the definition of the conformal barycenter in the Poincar\' e disc and argue that it is the natural concept of the mean (average). Calculation of this barycenter is one of key mathematical techniques implemented in layers of EmBolic.

Consider finite sets (configurations) of points in the Poincar\' e disc. Points are represented by complex numbers $z_1,\dots,z_N$, such that $|z_i|<1$. Introduce the following function
\begin{equation}
	\label{potential_disc}
	H(a) = - \sum_{i=1}^N \log \frac{(1-|a|^2)(1-|z_i|^2)}{|1-\bar a z_i|^2}, \quad a \in \mathbb{B}^2.
\end{equation}

It has been proven in \cite{Jaćimović_Kalaj_2025} that the function \eqref{potential_disc} has a unique minimum in ${\mathbb B}^2$. This minimum is defined as the conformal barycenter of the configuration $(z_1,\dots,z_N)$.

The conformal barycenter is conformally invariant, meaning that if $a$ is the conformal barycenter of points $\{z_1,\dots,z_N\}$, then for any $h \in \mathbb{G}_2$ the point $h(a)$ is the conformal barycenter of points $\{h(z_1),\dots,h(z_N)\}$. In other words, if the configuration is transformed by an isometry, then its barycenter is transformed by the same isometry.

In whole, minimum of the function \eqref{potential_disc} is the conformally invariant mean in the Poincar\' e disc. We refer to \cite{Jaćimović_Kalaj_2025} for all mathematical details and rigorous proofs.

Furthermore, one often deals with the configuration $\{z_1,\dots,z_N\}$ where certain weights $w_1,\dots,w_N$ are assigned to points. The definition of the weighted barycenter is introduced in \cite{Jaćimović_Kalaj_2025} with computational procedures for calculating (weighted) barycenters in ${\mathbb B}^2$ exposed in \cite{jacimovic2025group}.

\subsection{Random variate generation in the Poincar\' e disc}
\label{subsection_random_variate_generation}
Another technique relevant for EmBolic is sampling of random points in the Poincar\' e disc. Random sampling is necessary for the initialization of the optimization algorithm, but also for the conceptualization of the embedding method, as explained in Appendix~A.

We will sample random points from the probability distributions on $\mathbb{B}^2$ introduced in \cite{jacimovic2024conformally}. This family of probability distributions is defined by densities of the following form
\begin{equation}
	\label{conf-nat-disc}
	{\cal M}(z;a,s) = \frac{s-1}{\pi} \left( \frac{(1-|z|^2)(1-|a|^2)}{|1-\bar a z|^2} \right)^s.
\end{equation}
Parameters of the family are the mean (in hyperbolic metric) point $a \in {\mathbb B}^2$ and the concentration parameter $s>1$. 

As pointed out in \cite{jacimovic2024conformally}, all densities \eqref{conf-nat-disc} are unimodal and symmetric in hyperbolic metric. In addition, this family of distributions is closed with respect to actions of the group ${\mathbb G}$ of M\" obius transformations of the form \eqref{Mobius}.
We will refer to the family of probability distributions defined by densities \eqref{conf-nat-disc} as {\it M\" obius probability distributions in the Poincar\' e disc}. 

Another advantage of the M\" obius family is the simplicity of generation of a random sample. Suppose that we need to sample a random point in ${\mathbb B}^2$ with the mean $a=0$ and concentration parameter $s=s^*$. We start by sampling two uniformly distributed numbers on $[0,1]$. Denote them by $\gamma$ and $\kappa$ and set $\varphi = 2 \pi \gamma$ and $r = (1 - \sqrt[s^*-1]{1-\kappa})^2$. Clearly, $\varphi$ is uniformly distributed on $[0,2 \pi]$ and $r$ is a random number from the interval $(0,1)$. 

Now, we have a random point in ${\mathbb B}^2$ represented by a number $z=r e^{i \varphi}$. It is shown in \cite{jacimovic2024conformally} that the point $z$ is sampled from the probability distribution given by the density \eqref{conf-nat-disc} with parameters $a=0$ and $s=s^*$.

Now, due to the conformal-invariance of the family \eqref{conf-nat-disc} we can transform a random point $z$ to $\xi$ with $s=s^*$ and arbitrary mean point $a \in \mathbb{B}^2$ by acting on $z$ with the M\" obius transformation $g_a$, such that $g_a(0) = a$.

The above method provides a flexible way of generating random samples in ${\mathbb B}^2$ with the prescribed mean (Riemannian center of mass) and concentration. For higher values of $s>1$ we obtain a strongly concentrated sample around the mean point. On the other side, for lower values of $s$, the sample will be dispersed over the disc.

\section{Method}
\label{method}

In this Section we explain the EmBolic architecture step by step.

\subsection{Dataset}

We conduct experiments on a subset of the GoEmotions dataset \cite{demszky2020goemotions}, which consists of Reddit comments annotated with fine-grained emotion labels. We retain only the primary emotion label for each instance. After preprocessing and filtering, the dataset contains 4,884 instances annotated with 28 emotion categories (including neutral).

We use 4,744 instances for training and reserve 140 instances for testing, with five test examples randomly selected for each emotion to ensure balanced evaluation.

The emotion set includes \emph{admiration, amusement, anger, annoyance, approval, caring, confusion, curiosity, desire, disappointment, disapproval, disgust, embarrassment, excitement, fear, gratitude, grief, joy, love, nervousness, optimism, pride, realization, relief, remorse, sadness, surprise}, and \emph{neutral}.

\subsection{Preprocessing}

We first tokenize and lemmatize the text, normalize URLs, hashtags, and emojis, and remove stopwords while retaining negations and intensifiers (e.g. "not", "very"). From the preprocessed text, we construct a word-emotion co-occurrence matrix \(M \in \mathbb{R}^{|V| \times 28}\), where \(|V|\) is the vocabulary size and \(28\) represents the number of emotion categories. Each entry \(M_{w,e}\) represents the normalized frequency with which word \(w\) co-occurs with emotion \(e\).

We then compute a word-word similarity matrix \(S \in [0,1]^{|V| \times |V|}\) using a scaled Euclidean distance by 1/2 to ensure \(S_{ij} \in [0,1]\):
\begin{equation*}
	S_{i,j} = 1 - \frac{1}{2} \sum_{e=1}^{28} \left( M_{i,e} - M_{j,e} \right)^2,
\end{equation*}
so that words associated with similar emotions have higher similarity scores. This matrix is used as input to Hyperbolic GloVe to learn emotion-informed word representations.

\subsection{Hyperbolic GloVe}
\label{sec:hyperbolic-glove}

Hyperbolic GloVe tends to group together words based on their associations with specific emotions, thus providing meaningful, emotion-informed representations. Messages are then represented as sequences of words, that is - as ordered configurations of points in the Poincar\' e disc. A mathematical background for Hyperbolic GloVe is probabilistic matrix factorization in hyperbolic space, explained in Appendix~A. This procedure is analogous to the standard (Euclidean) probabilistic matrix factorization \cite{mnih2007probabilistic}, although employing different geometric and statistical apparatus.

\subsection{Emotion attention}

In order to obtain message representations that reflect emotional content, we implement a kind of emotion attention mechanism based on the rigorous mathematical apparatus presented in Section \ref{mathematical_preliminaries} and built on the top of the Hyperbolic GloVe embeddings. Each message is represented as an ordered configuration of points in the Poincar\' e disc. The goal of the attention mechanism is to learn which (combinations of) words contribute most to the message emotional meaning.

EmBolic does not assign fixed weights to words. Instead it computes the attention weights from the entire configuration of points. This provides the model with the necessary flexibility. To that end, the complex-valued word embeddings are converted into a single real vector. This vector is then passed through a learnable linear layer, followed by the softmax, producing a set of attention weights that sum to one.

These weights are further used to compute the weighted conformal barycenter (as defined in subsection \ref{subsection_conformal_barycenter}) of word embeddings. This pooling procedure yields a single point in the Poincar\' e disc that serves as representation of the message.

The parameters learned by EmBolic are weights of the linear projection, rather than attention values themselves. This allows the model to flexibly assign different attention patterns to different messages.

\begin{figure}[t]
	\begin{center}
		\includegraphics[width=1.0\textwidth]{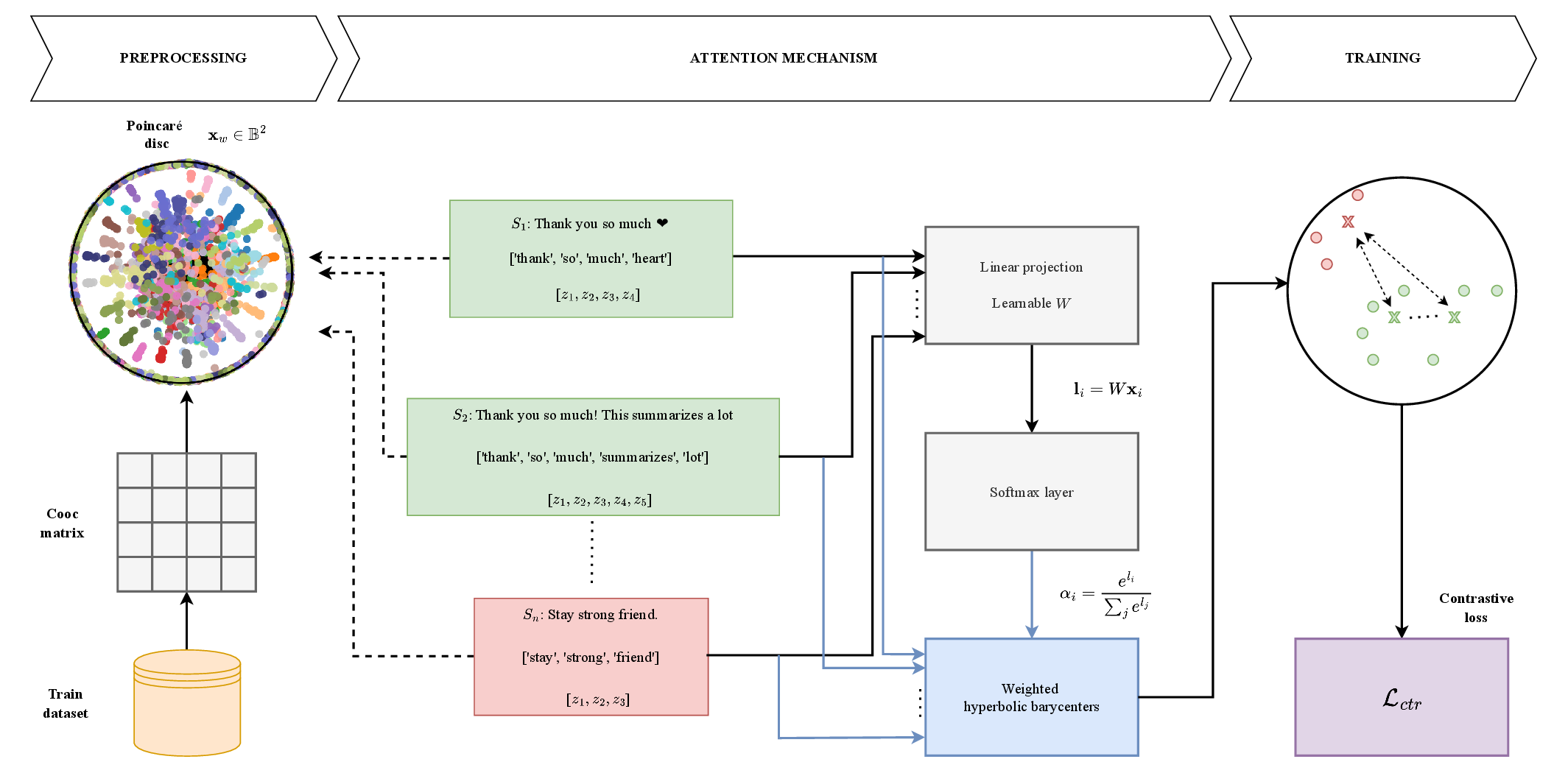}
	\end{center}
	\caption{Embolic architecture.}
	\label{fig3}
\end{figure}

The full EmBolic architecture is schematically presented in Figure \ref{fig3}.

\subsection{Loss function}

The model is trained on the loss function consisting of three terms: positive sampling, negative sampling and regularization. 

\begin{equation}
	\label{loss}
	\mathcal{L}_{ctr}
	=
	\frac{1}{|\mathcal{P}|}
	\sum_{(i,j)\in\mathcal{P}}
	d_{h}(\mathbf{b}_i,\mathbf{b}_j)
	-
	\frac{1}{|\mathcal{N}|}
	\sum_{(i,j)\in\mathcal{N}}
	d_{h}(\mathbf{b}_i,\mathbf{b}_j)
	-
	\lambda
	\frac{1}{B}
	\sum_{i=1}^{B}
	\left[\log\!\left(1-\|\mathbf{b}_i\|^2\right)\right]
\end{equation}
where $b_i$ are computed conformal barycenters representing messages.

Objective function \eqref{loss} defines the contrastive loss. The first term pulls barycenters corresponding to the same emotion closer together, while the second term pushes barycenters corresponding to different emotions further apart. Finally the third term prevents barycenters from getting too close to the boundary circle or escaping outside the disc. This is an important ingredient as otherwise the model may find out that is beneficial to make certain distances infinitely high.

Parameters of the loss function are carefully tuned in such a way to achieve the balance between positive and negative samples, as well as optimal utilization of the embedding disc space.

\section{Results}

\label{results}

We train our model on 4,744 instances from the GoEmotions dataset and validate the model on 140 previously unseen instances (5 messages for each emotion).

\subsection{Word embeddings}

We perform three independent word embeddings in the Poincar\' e disc using Hyperbolic GloVe. In such a way the representation space is the 3-disc (a six-dimensional manifold).

\begin{figure}[h!]
	\centering
	\includegraphics[width=1.0\columnwidth,height=0.3\textheight,keepaspectratio]{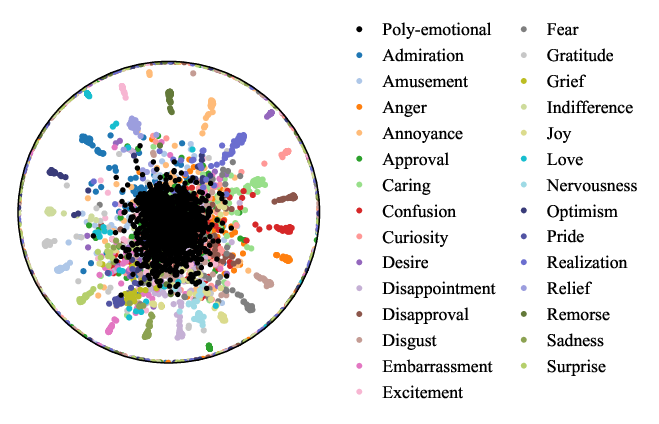}
	\caption{Hyperbolic GloVe word embeddings in the Poincar\'e disc.}
	\label{fig4}
\end{figure}

Figure \ref{fig4} illustrates word embeddings in one disc.

Each point represents one word. Words which are significantly correlated with a specific emotion are depicted in the corresponding color, while poly-emotional words are shown in black. Somehow unexpectedly we immediately observe almost monochromatic rays. This indicates that the optimization procedure tends to associate emotions with the directions in the plane. On the other hand, poly-emotional words (black dots) tend to group together around the center of the disc, thus encoding their semantic ambiguity.

\begin{figure}[h]
	\centering
	\includegraphics[width=0.5\columnwidth, height=0.4\textwidth, keepaspectratio]{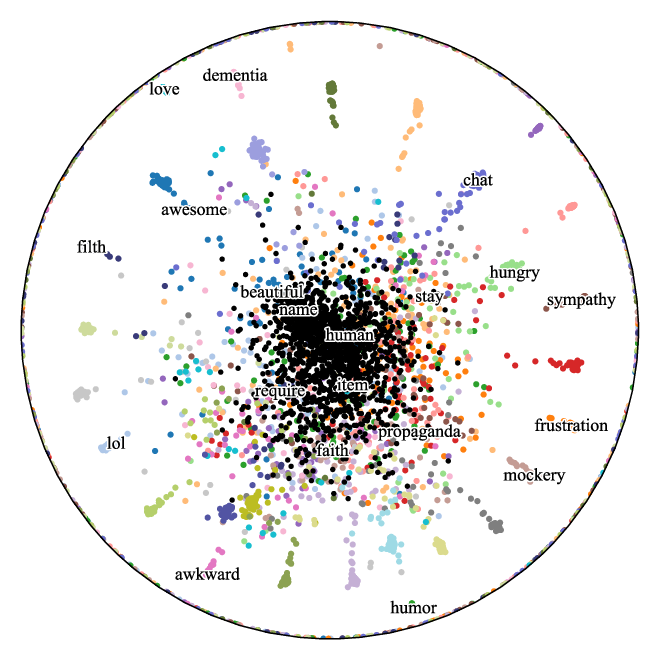}
	
	\caption{Randomly selected words with their representations in the Poincar\'e disc. This illustrates emotion-related semantic ambiguity of certain words.}
	\label{fig5}
\end{figure}

Figure \ref{fig5} shows a random subset of words illustrating how hyperbolic metric captures hierarchical relationships. In particular, those words which are not strongly associated with any specific context are embedded near the center of the disc.

\subsection{Text embeddings}
\label{subsection_text_embeddings}
\begin{figure}[t]
	\centering
	\includegraphics[width=0.32\textwidth]{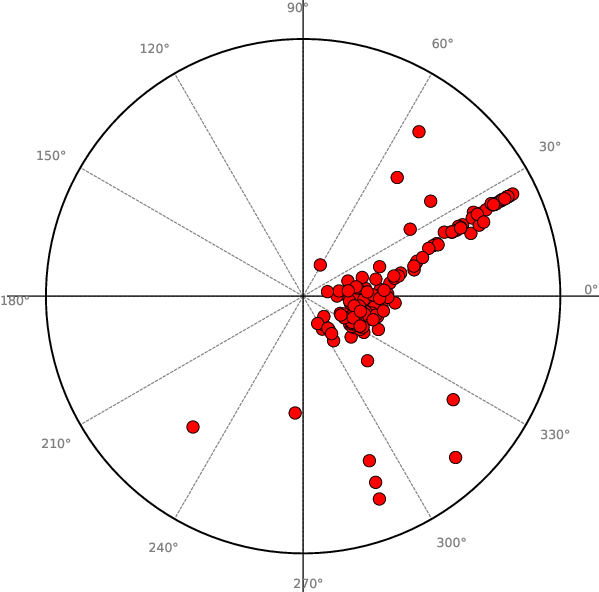}
	\hfil
	\includegraphics[width=0.32\textwidth]{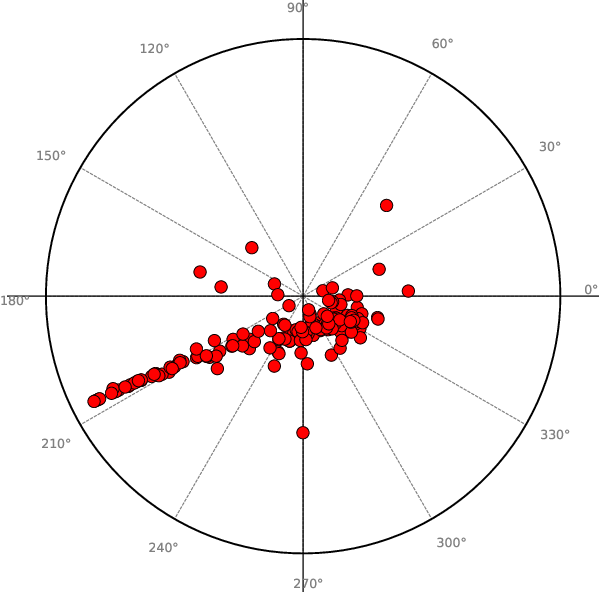}
	\hfil
	\includegraphics[width=0.32\textwidth]{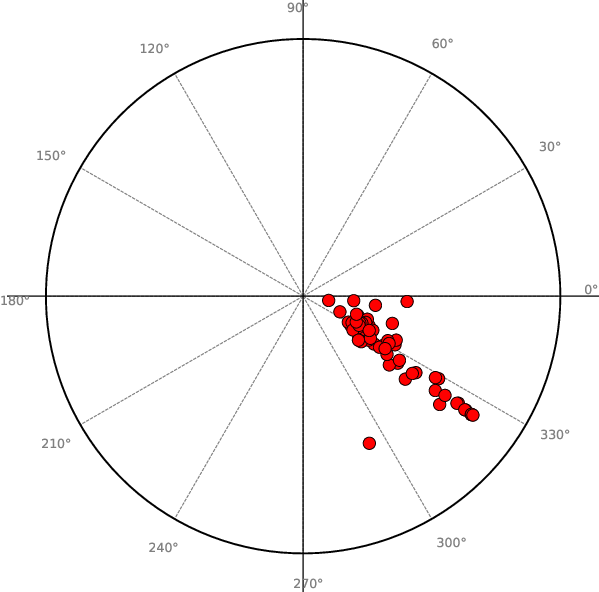}
	
	\vspace{2mm}
	
	\includegraphics[width=0.32\textwidth]{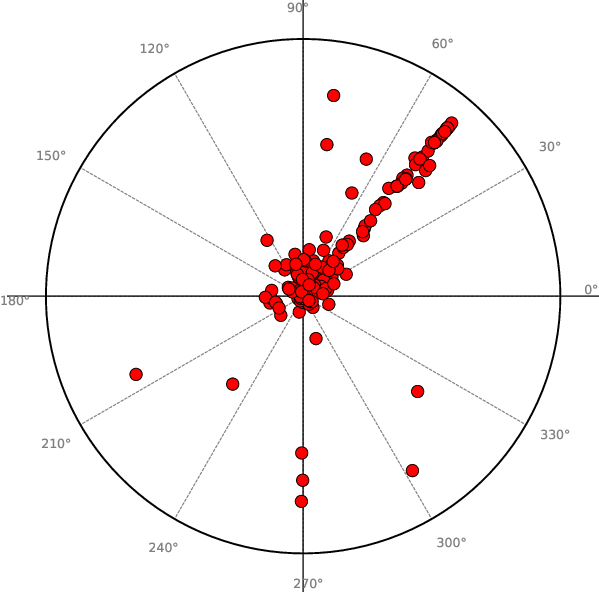}
	\hfil
	\includegraphics[width=0.32\textwidth]{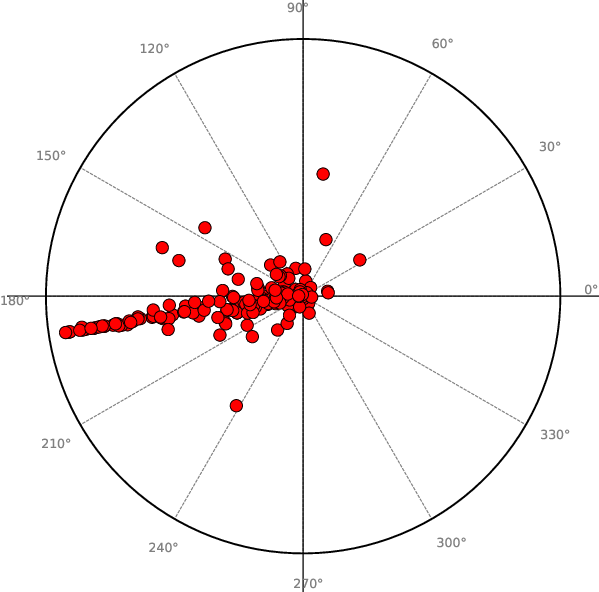}
	\hfil
	\includegraphics[width=0.32\textwidth]{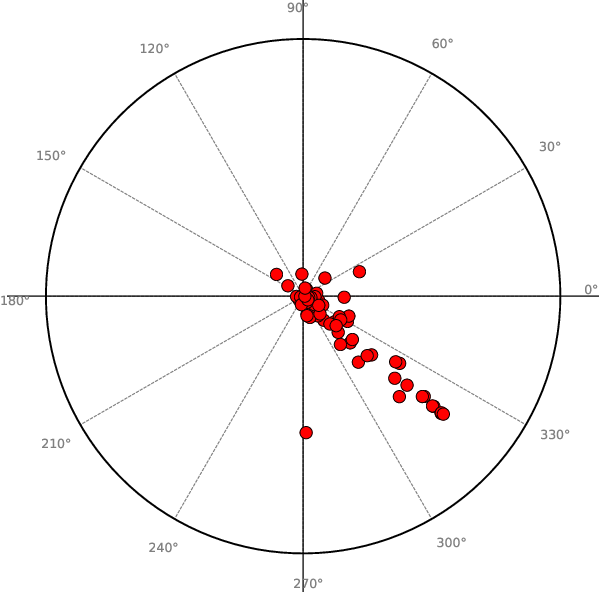}
	
	\caption{Learned representations of textual messages for Desire, Love, and Nervousness (top row) and the corresponding representations after isometric (M\"obius) correction (bottom row).}
	\label{fig6}
\end{figure}

By minimizing the loss function \eqref{loss} the model seeks to group instances (messages) from the training set into separate clusters based on their annotations. By computing the conformal barycenters for each message the model learns their representations as points in the disc. Embeddings of instances annotated by three emotions in the disc are depicted in Figure \ref{fig6}.

We immediately notice that representations of messages annotated by the same emotion form a finger-like structure. The "fingers" point to different directions. Emphasize that such directional structure was not imposed by any modeling assumption. This means that the model by itself suggests circular (or toroidal, or spherical) geometry, where each emotion corresponds to a direction (or several directions) in the plane (or in the higher-dimensional space).

We further observe that majority of instances are densely packed around the common epicenter, located to the south-west from the center of the disc. Although Figure \ref{fig6}, upper row shows only three emotions, this epicenter is the same for all 28 emotions. Obviously, instances which correspond to points in this epicenter can not be successfully clusterized. However, those instances which are embedded along the fingers, can be separated into well-defined clusters. Obviously, this can be achieved by associating each emotion with a certain direction in the plane. However, one can notice that fingers are not well aligned with the radii. In order to correct this, the model further learns a M\" obius transformation \eqref{Mobius}, which maps the epicenter to the center of the disc. Since \eqref{Mobius} are isometries of the Poincar\' e disc, this transformation fully preserves the information, as long as the same transformation is applied to all fingers. After this correction, fingers get well aligned along the radii, as shown in Figure \ref{fig6}, bottom row. This renders almost perfect correspondence between emotions and angles. These angles correspond to dominant directions, calculated as the weighted average, where the weight of each point is proportional to the square of its modulus 
\begin{equation*}
	\psi_j = \sum \limits_{k=1}^m r_k^2 \varphi_k.
\end{equation*}
Here, $m$ is the number of instances for each emotion and $z_k = r_k e^{i \varphi_k}$ are representation points corresponding to that emotion.

As explained above, we obtain one-to-one mapping between emotions and angles. Underline that different embeddings yield different configurations of emotions on the circle. However, multiple simulations unveil strong statistical correlations between emotions. By performing embeddings in $k$ Poincar\' e discs, we finally obtain emotion representations by $k$ angles, e.g. by a point on the $k$-torus. (Again, in the present study we performed three embeddings, i.e. $k=3$.)

Overall, the model does not confirm Assumption A2 from subsection \ref{subsection_modeling_assumption}. Instead, it suggests that the space of emotions might exhibit positive curvature (toroidal or spherical, rather than hyperbolic geometry). On the other hand, hyperbolic geometry allows for a subtle assessment, by encoding semantic uncertainties arising when trying to relate words with emotions.

A complete visualization for one embedding (28 fingers) is presented in the Supplemental Material. These figures confirm that each finger points to a different direction with emotions almost uniformly distributed on the circle. This enables for an efficient utilization of the representation space and separation of emotions.

\subsection{Busemann energy scores}
\label{validation_scores_accuracy}
Once the model is trained, each message is represented by a point $z_k$ in the Poincar\' e disc, while emotions are represented by directions $\xi_j = e^{i \psi_j}, \; j=1,\dots,28$. In order to predict emotions associated with (previously unseen) instances, we generate their queries and evaluate their alignment with keys (class directions). This alignment will be measured by the Busemann energy function:
\begin{equation}
	\label{Busemann}  
	B_{\xi_j}(z_k) = \log \frac{1-|z_k|^2}{1 - \bar z_k \xi_j}, \quad z_k \in {\mathbb B}^2, \; \xi_j \in {\mathbb S}^1.
\end{equation}
Exponentiating the Busemann function we obtain an analogy of the Boltzmann weights along directions:
\begin{equation}
	\label{Poisson_kernel}
	p_{z_k}(\xi_j) = \frac{1-|r_k|^2}{1 - 2 r_k \cos(\psi_j - \Phi_k) + r_k^2},
\end{equation} 
where $\xi_j = e^{i \psi_j}$ and $z_k = r_k e^{i \Phi_k}$.

Evaluations of \eqref{Poisson_kernel} yield 28 scores for each message. These scores are interpreted as (exponents of) energies between queries and keys.

\begin{remark}
	The functions of the form \eqref{Poisson_kernel} are well-known as Poisson kernels in complex analysis. In directional statistics, (multiplied by the normalizing factor $1/2 \pi$) they appear as densities of the wrapped Cauchy distributions on the circle \cite{mccullagh1996mobius}. These functions ensure the scoring system that is well aligned with the geometry of the representation space. Indeed, $\Phi_k$ is an angle of the embedded message and higher scores will be assigned to those emotions represented by angles $\psi_j$ which are closer to $\Phi_k$. On the other side, radius $r_k$ tells us how confident the model is about its prediction. For values $r_k$ close to one, the best scores will be very high, indicating strong
	confidence. In the case when $r_k = 0$, the Busemann energy \eqref{Busemann} is equal to zero and the function \eqref{Poisson_kernel} is constant. This means that the model assigns equal scores to all emotions, indicating total uncertainty. On the other extreme the limit case with $r_k = 1$ and $\Phi_k = \psi_j$ corresponds to the delta distribution, assigning an infinite
	score to emotion $j$. This implies that the model is absolutely confident about the prediction. Notice that the loss function ensures that all representations are located strictly in the interior of the unit disc, i.e. $r_k < 1$. Hence, the scoring system naturally quantifies the confidence level and allows for at least a low degree of uncertainty in each prediction. 
\end{remark}

We have explained the scoring system for embeddings in a single Poincar\' e disc. For each message and each disc we obtain 28 scores, where each score is a positive real number. Then the total score for each emotion is an average of scores for each disc. \footnote{Emphasize the possibility of assigning weights to discs, meaning that we take a weighted average of scores. In general, this may significantly improve the performance of the model, but requires an additional training stage for learning optimal weights.}

\begin{figure}[h]
	\centering
	\includegraphics[width=0.32\textwidth]{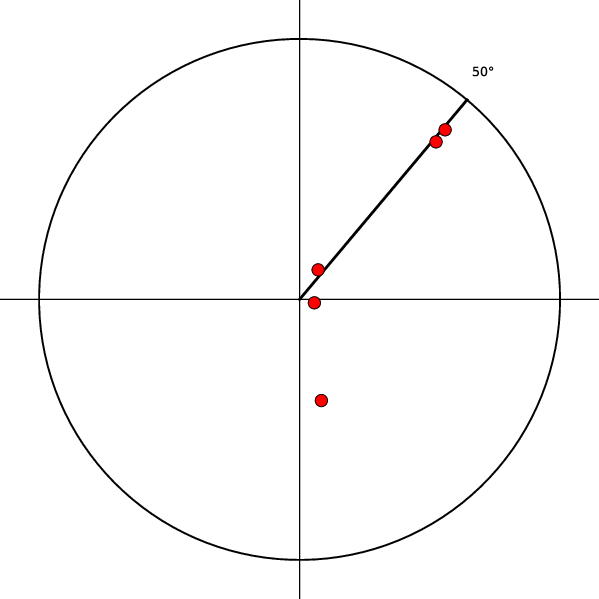}
	\hfil
	\includegraphics[width=0.32\textwidth]{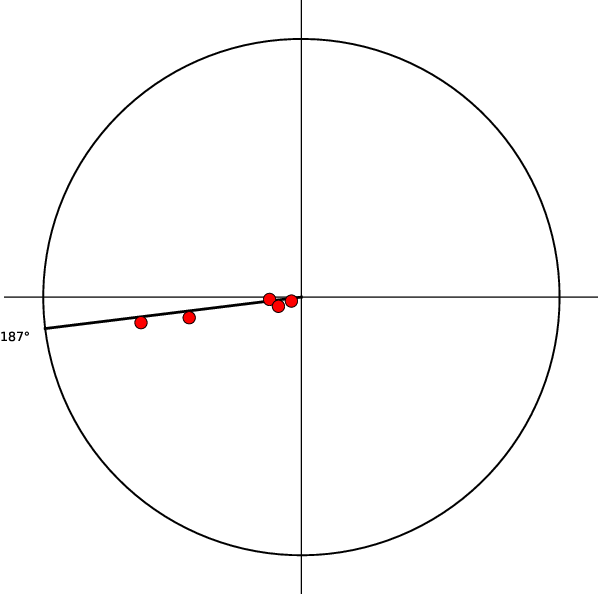}
	\hfil
	\includegraphics[width=0.32\textwidth]{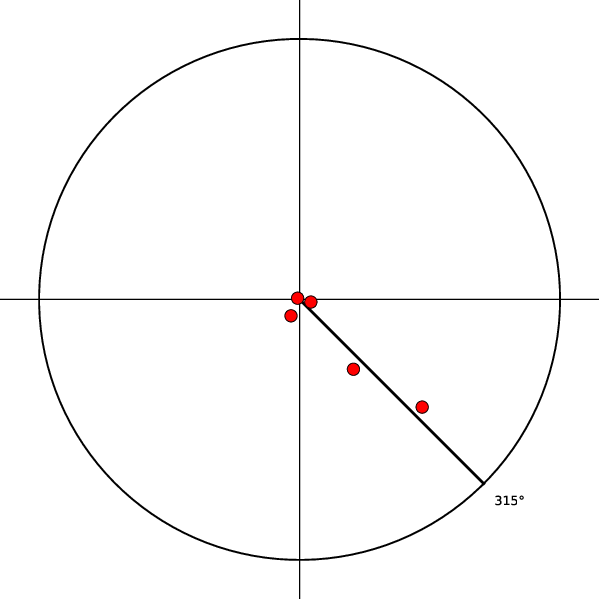}
	\caption{Visualization of the validation results. Representations of test	messages for Desire, Love and Nervousness and the radii encoding these emotions. Correct predictions correspond to the points which are placed on the radii. The confidence levels are encoded by the moduli (distances from the center).
	}
	\label{fig7}
\end{figure}

In Figure \ref{fig7} we visualize embeddings of previously unseen instances along with the directions representing true emotions for each instance. This illustrates that higher scores (implying a stronger confidence) are obtained for those embeddings where points are closer to the boundary circle. For instance, the left panel demonstrates two correct highly confident predictions; two predictions with low confidence and one highly confident misprediction for Desire. The middle panel demonstrates all five correct predictions for Love, although three of them with low confidence level. In the Supplemental Material we provide more details with figures for all instances from the test set and for all emotions, for the embedding in one disc.

Finally, 28 scores can be transformed into a sequence of numbers that can be interpreted as probabilities. For that goal, we pass the scores through the softmax with the temperature equal to $T=0.05$, which is manually (and to a certain point arbitrarily) chosen.

We report the classification performance of our model using top-1, top-3, and top-5 accuracy metrics.

Accuracy results for three individual simulations are given in Table \ref{tab1}. 

\begin{table}[h]
	\centering
	\small
	\caption{Accuracy results for individual embeddings}
	\label{tab1}
	\begin{tabular}{lcccc}
		\toprule
		Simulation & Total samples & Top-1 & Top-3 & Top-5 \\
		\midrule
		Simulation 1 & 140 & 0.150 & 0.286 & 0.414 \\
		Simulation 2 & 140 & 0.143 & 0.343 & 0.457 \\
		Simulation 3 & 140 & 0.186 & 0.307 & 0.429 \\
		\bottomrule
	\end{tabular}
\end{table}

Cumulative accuracy report for the 3-disc is substantiated in Table \ref{tab2}. While the total accuracy after three embeddings is approximately 26 \%, the true emotion listed in top-5 is 58 \%, indicating the potential for drastic improvements with a moderate increase of the dimension.

\begin{table}[h]
	\centering
	\small
	\caption{Cumulative accuracy results}
	\label{tab2}
	\begin{tabular}{lccc}
		\toprule
		Total samples & Top-1 & Top-3 & Top-5 \\
		\midrule
		140 & 0.264 & 0.464 & 0.579 \\
		\bottomrule
	\end{tabular}
\end{table}

\subsection{Error analysis and confidence assessment}

In the present subsection we conduct a brief confidence assessment and error analysis. We examine some instances which are misclassified with a high confidence level, as well as several instances about which the model remained highly indecisive.

By setting the confidence level at 20\% we obtain the accuracy statistics shown in Table \ref{tab3}. Notice that the model provides a confident prediction for only less than quarter of instances. Among these instances, the prediction is correct in nearly 55 percents of cases.

\begin{table}[h]
	\centering
	\small
	\caption{Confidence analysis of model predictions with a certainty threshold of 20\%}
	\label{tab3}
	\begin{tabular}{lcc}
		\toprule
		Category & Count & Percentage (\%) \\
		\midrule
		Secure correct (Top-1 True, $\geq$ 20\%) & 17 & 12.14 \\
		Secure wrong (Top-1 False, $\geq$ 20\%) & 15 & 10.71 \\
		Insecure ($<$ 20\%) & 108 & 77.14 \\
		\midrule
		Total & 140 & 100.00 \\
		\bottomrule
	\end{tabular}
\end{table}

For the better understanding of misclassifications and uncertainties, we start with five representative messages shown in Table \ref{tab4}. 

\begin{table}[h!]
	\centering
	\small
	\caption{High-confidence mispredictions of the emotion classification model}
	\label{tab4}
	\begin{tabular}{p{7cm} l l c l}
		\toprule
		Message & True label & Top prediction & Score (\%) & True rank \\
		\midrule
		I only use freedom units, sorry. & Approval & Remorse & 75.41 & -- \\
		I felt the same guilt when I stopped believing in [NAME]. & Remorse & Caring & 75.60 & -- \\
		Ahhhh fond memories. & Joy & Sadness & 53.94 & Top-5 (4.83) \\
		Yes, I've noticed it too. Or the photo of the item won't load. This company is so shameful. & Embarrassment & Confusion & 48.63 & Top-2 (42.54) \\
		Thanks for the laugh lol. & Gratitude & Amusement & 33.13 & Top-2 (13.04) \\
		\bottomrule
	\end{tabular}
\end{table}

The first two sentences are misclassified with a very high level of confidence. Clearly, in the first example the prediction Remorse (instead of Approval) is triggered by the word "sorry". In the second sentence, the "guilt for stopping believing in..." is interpreted as an empathy for others rather than a remorse.

The predicted emotion (with a high confidence level) for the third sentence is Sadness whereas the true label is Joy. This is easily explained by the ambivalence of this sentence with a sigh that can be interpreted as an expression both of sadness and joy. 

For the fourth message, the model is hesitant between Embarrassment and Confusion (the true label), which seems as an adequate prediction. The same applies to the fifth sentence, where the model predicts Amusement, while the true label (Gratitude) is listed as the second choice. The slang word "lol" was understood as amusement in this example.

Table \ref{tab5} contains three examples where the model failed to guess the true label, but listed it among top-5 predictions. One can see that predictions seem quite reasonable. Notice the emoji in the second sentence that was possibly interpreted as an embarrassment or a teasing tone.

\begin{table}[h!]
	\centering
	\small
	\caption{Examples where the true label appears in the top-5 predictions}
	\label{tab5}
	\begin{tabular}{p{7cm} l l c l}
		\toprule
		Message & True label & Top prediction & Score (\%) & True rank \\
		\midrule
		What a beautiful message—it’s all about love, and I hope it sinks in to some of the COB people who might see it. & Optimism & Admiration & 21.05 & Top-5 (4.52) \\
		Sorry my teams gave you your only two losses this season ;) & Remorse & Embarrassment & 25.85 & Top-2 (22.78) \\
		Here for [NAME], love this. & Love & Excitement & 36.10 & Top-2 (34.47) \\
		\bottomrule
	\end{tabular}
\end{table}

Finally, Table \ref{tab6} contains two examples for which the model showed low confidence levels.

\begin{table}[h!]
	\centering
	\small
	\caption{Examples of uncertain model predictions}
	\label{tab6}
	\begin{tabular}{p{7cm} l l c l}
		\toprule
		Message & True label & Top prediction & Score (\%) & True rank \\
		\midrule
		
		[NAME] will be gone. Although I think end of the first is where he should go, he will go super early. & Optimism & Admiration & 4.69 & -- \\
		This is so scary. Keep safe! I would put fencing up and become a licensed carrier at that point. Sounds like crazy druggies! & Nervousness & Nervousness & 5.99 & Top-1 (5.99) \\
		\bottomrule
	\end{tabular}
\end{table}

Notice that in the second example the true label is correctly predicted, although with the confidence level of only 6 \%.  Overall error analysis unveils potential for capturing semantic ambiguities. In fact, the model provides meaningful predictions even in some of those cases where it fails to predict the true label.

\subsection{Emotion maps}

\begin{figure}[h!]
	\centering
	\includegraphics[width=\columnwidth, height=0.5\textwidth, keepaspectratio]{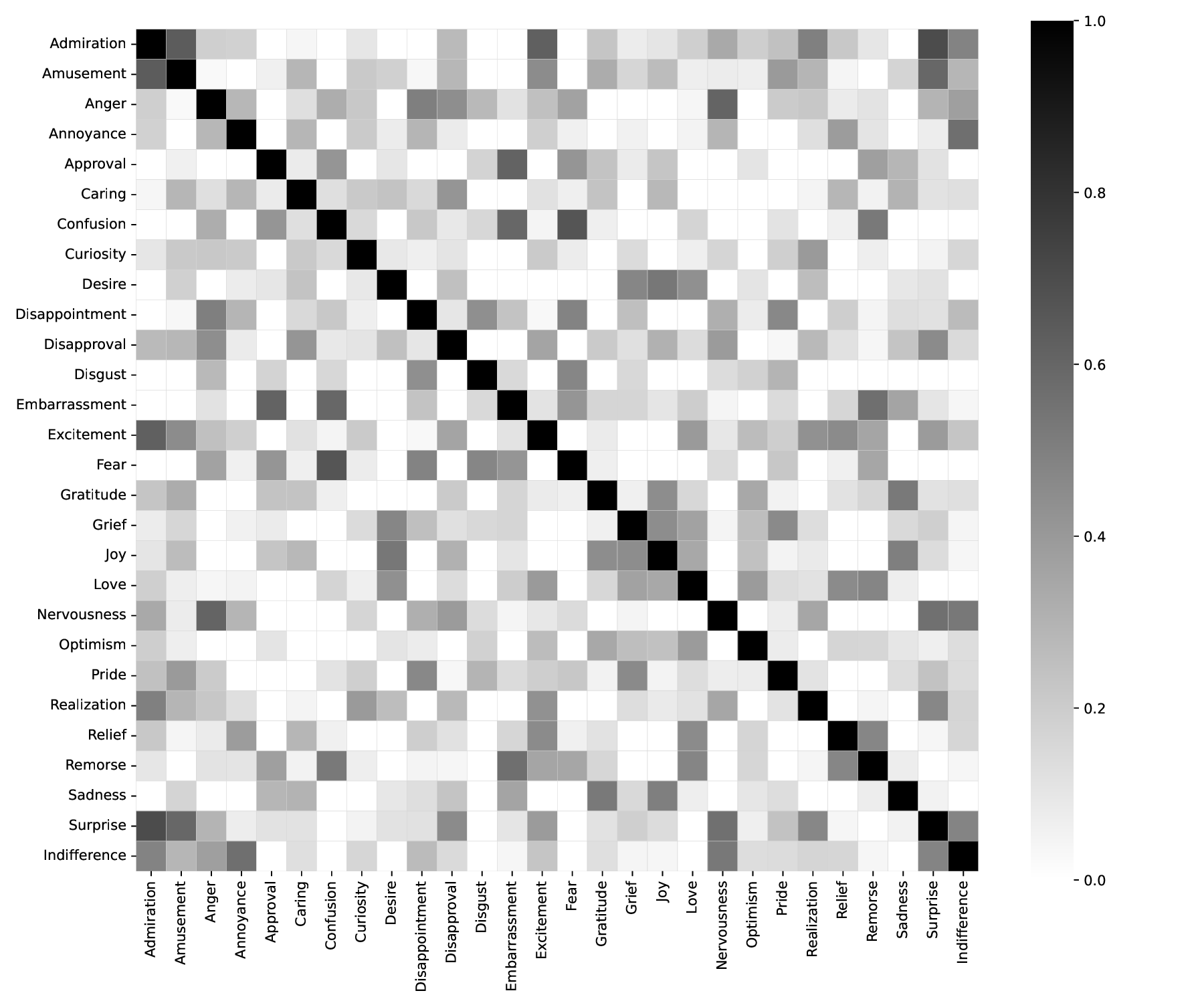}
	\caption{ Emotion correlation matrix learned by EmBolic. Each cell $(i, j)$ show how frequently emotion $i$ appears together with $j$ in top 5. Darker cells indicate stronger pairwise correlations.}
	\label{fig8}
\end{figure}

Based on predictions for instances from the test set, we construct the co-occurrence matrix to visualize correlation patterns on the set of emotions. Figure \ref{fig8} presents the matrix which shows how frequently each pair of emotions co-occur in top-5 predictions.

By embedding this matrix into the Poincar\' e disc using Hyperbolic GloVe, we obtain the emotion map depicted in Figure \ref{fig9}. This map shows distances between emotions based on their co-occurrences and relations with textual messages. Emphasize that each embedding of this matrix yields different emotion maps, where certain clusters appear and vanish. This indicates that the two-dimensional disc is insufficient for mapping emotions. Instead, the emotion map gets more faithful if emotions are represented as points in the multi-disc.

\begin{figure}[h]
	\centering
	\includegraphics[width=\columnwidth, height=0.4\textwidth, keepaspectratio]{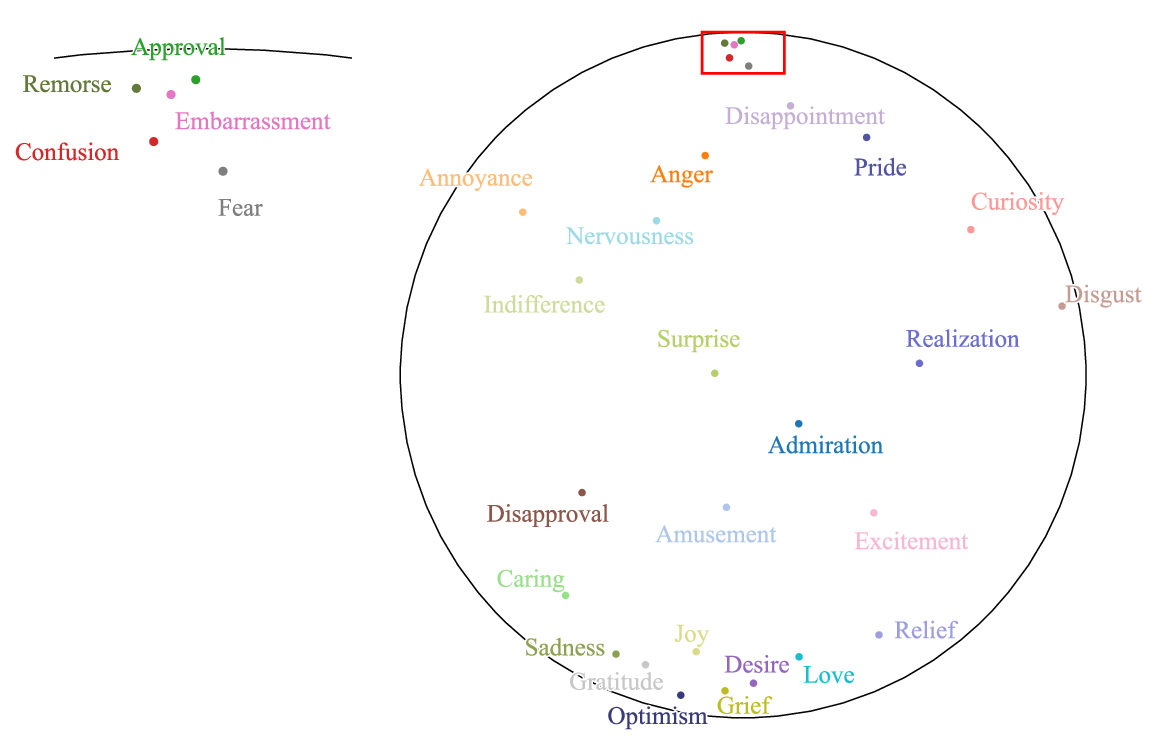}
	\caption{Map of emotions in the Poincar\' e disc learned from the co-occurrence matrix of emotions depicted in Figure \ref{fig8}.}
	\label{fig9}
\end{figure}

\section{Conclusion}
\label{conclusion}
We presented a hyperbolic DL architecture for fine-grained emotion analysis named EmBolic. Our study supports previous results on hyperbolic representations in language models and demonstrates their potential in Emotion AI.

It is frequently claimed in the literature that advantages of hyperbolic representations are due to the exponential growth of the volume with respect to the radius. This property has also been rephrased by stating that "the hyperbolic disc is a continuous version of the tree." \cite{ghys2006poincare}

Our study unveils a slightly subtler explanation of advantages of hyperbolic ML. Contexts are naturally represented by directions in the space, which implies spherical, or toroidal geometry. From this point of view, hyperbolic geometry faithfully captures uncertainty, or semantic ambiguity. Put differently, we effectively assume that words are represented by probability distributions on spheres or tori, with the contextual uncertainty encoded by points (conformal barycenters of probability distributions) in hyperbolic balls or discs. Common words are represented by the (almost) uniform distribution, with the conformal barycenter close to the center of the hyperbolic disc. In contrast, specific words that imply a narrow context correspond to the strongly peaked distributions and their hyperbolic representations are points far from the center. Therefore, EmBolic learns {\it hierarchy of uncertainties} on the set of lexical sentences, quantified by Busemann energies \eqref{Busemann}. \footnote{The idea of using the Busemann function for attention patterns in hyperbolic spaces has been explored in several recent studies \cite{chen2026hyperbolicbusemannneuralnetworks,ghadimi2021hyperbolic}.}

Our results also emphasize scalability of hyperbolic models with respect to the dimension of the representation space. In our case this means that repeated embeddings extract new information. While each single embedding in the Poincar\' e disc has a modest accuracy, combining several embeddings yields consistent improvements. In the present study, we reported results for the model trained on five thousand messages from the GoEmotions dataset, with the dimension of representation space equal to six (product of three Poincar\' e discs). There are straightforward ways of achieving higher accuracy: a) increasing the dimension (number of discs); b) training the model on larger dataset and c) adding an additional small neural network for adjusting weights of scores for each disc.

Our model may apply to various tasks and datasets beyond Emotion AI where semantic ambiguity appears as a key challenge. Essentially the same architecture may prove efficient in inferring political views from narratives, or diseases from DNA sequences.

\section*{Acknowledgments}
	V. Ja\' cimovi\' c acknowledges the support of the Ministry of Education, Science and Innovation of Montenegro (projects "Mathematical analysis, optimization and machine learning" and "Complex-analytic and geometric techniques for non-Euclidean ML: theory and practice"). 
	
	Both authors acknowledge the support from the NetApp Faculty Fellowship fund, project title "Hyperbolic text embedding". 

\appendix

\section{Hyperbolic Matrix Factorization}
\renewcommand{\theequation}{A.\arabic{equation}}
\renewcommand{\thefigure}{A.\arabic{figure}}
\renewcommand{\thetable}{A.\arabic{table}}

\renewcommand{\theHequation}{A.\arabic{equation}}
\renewcommand{\theHfigure}{A.\arabic{figure}}
\renewcommand{\theHtable}{A.\arabic{table}}

\setcounter{equation}{0}
\setcounter{figure}{0}
\setcounter{table}{0}
Initial embeddings of words are obtained as a result of the optimization procedure which we refer to as Hyperbolic GloVe. This procedure is based on the novel matrix factorization method in the Poincar\' e disc. In this Appendix we briefly explain this method.

We follow mathematical reasoning analogous to the one in standard probabilistic matrix factorization model proposed by \cite{mnih2007probabilistic}, but with an adaptation of the statistical apparatus to hyperbolic geometry. Most notably, we will assume M\" obius (instead of Gaussian) priors (\ref{conf-nat-disc}) on word feature points in ${\mathbb B}^2$.

As explained in subsection \ref{sec:hyperbolic-glove} the word co-occurence matrix $S =\{ s_{ij} \}$ contains information on relationships between words and emotions. This matrix is normalized in such a way to contain entries between zero and one. Assume that $u_i$ are feature points of words in the Poincar\' e disc ${\mathbb B}^2$ and denote by $U$ the whole set of points (complex numbers) $u_1,\dots,u_N$.

Introduce the function $g(x) = 2 e^{-\alpha x}/(1 + e^{-\alpha x})$ and denote $w(u_i,u_j) = g(d_h(u_i,u_j))$, where $d_h(\cdot,\cdot)$ denotes the hyperbolic distance between points $u_i$ and $u_j$. Here, $\alpha>0$ is the hyperparameter of the algorithm. 

We assume mutually independent Gaussian priors on the matrix entries:
\begin{equation*}
	p(S|U,\sigma^2) = \prod_{i=1}^N \prod_{j=1}^N {\cal N}\left(s_{ij}|w(u_i,u_j),\sigma^2\right)
\end{equation*}

As feature points lie in the hyperbolic disc, rather than in Euclidean space, the Gaussian family of distributions is inadequate. Instead, we place M\" obius priors (\ref{conf-nat-disc}) on word feature points in ${\mathbb B}^2$. We set zero mean (barycenter of the probability distribution) $a=0$:
\begin{equation}
	\label{Moebius_prior}
	p(U|s) = \prod_{i=1}^N {\cal M} (u_i|0,s)
\end{equation}

The log of the posterior distribution for feature points reads
\begin{align}
	\label{log_posterior}
	\ln p(U|S,\sigma,s) &= - \frac{1}{2 \sigma^2} \sum_{i=1}^N \sum_{j=1}^N (c_{ij}-w(u_i,u_j))^2
	+ s \sum_{i=1}^N \ln (1-|u_i|^2) - N^2 \ln \sigma \nonumber \\
	&\quad   - \frac{1}{2} N^2 \ln(2 \pi) + N \ln (s-1) - N \ln \pi
\end{align}

Discarding constants and rearranging multipliers we recast the maximization problem for the function \eqref{log_posterior} as the following minimization problem
\begin{align}
	\label{objective}
	\mbox{Minimize } E(u_1,\dots,u_N) &= 
	\sum_{i=1}^N \sum_{j=1}^N (s_{ij}-w(u_i,u_j))^2 - \lambda \sum_{i=1}^N \ln (1-|u_i|^2)
\end{align}
with respect to $u_i \in {\mathbb B}^2$.

Now, it is clear that M\" obius priors \eqref{Moebius_prior} effectively result in the regularization term $\ln(1-|u_i|^2)$ in the objective function \eqref{objective}. This term appears instead of the Frobenius norm in the standard probabilistic matrix factorization model \cite{mnih2007probabilistic} and plays approximately the same role. It penalizes large pairwise distances, with penalizing multiplier $\lambda>0$. In other words, regularization prevents approaching points too closely to the boundary circle. On the other hand, a very strong regularization may result in stretching almost all points inside the smaller sub-disc in ${\mathbb B}^2$ thus preventing utilization of the whole embedding space. These challenges are alleviated by fine-tuning hyperparameters $\lambda$ and $\alpha$ in order to avoid numerical instabilities and utilize the whole disc at the same time.

Feature points (embeddings) of the words are obtained by applying the gradient descent algorithm in the Poincar\' e disc. This procedure deploys an adaptive step respecting hyperbolic metric (\ref{tensor}) in ${\mathbb B}^2$.

In order to initialize the gradient descent we start with a random sample generated from the distribution (\ref{conf-nat-disc}), as explained in subsection \ref{subsection_random_variate_generation}. Empirical result suggest higher values for the concentration parameter $s$ (say $s \approx 10$), resulting in a highly concentrated initial distribution of points. Given a fine balance between positive and negative samples, the gradient descent procedure pretty efficiently redistributes these points over the disc filling in almost the whole disc.

\section{Visualization of embeddings for all 28 emotions}
\renewcommand{\theequation}{B.\arabic{equation}}
\renewcommand{\thefigure}{B.\arabic{figure}}
\renewcommand{\thetable}{B.\arabic{table}}

\renewcommand{\theHequation}{B.\arabic{equation}}
\renewcommand{\theHfigure}{B.\arabic{figure}}
\renewcommand{\theHtable}{B.\arabic{table}}

\setcounter{equation}{0}
\setcounter{figure}{0}
\setcounter{table}{0}

For the sake of completeness of the exposition, we supplement our paper with additional figures visualizing embeddings of words and messages. 

Figure \ref{fig4} in the main text shows word embeddings in the disc learned in one simulation. 

Figure \ref{figB1} presents word embeddings obtained in two additional simulations.

\begin{figure}[h]    
	\centering
	\includegraphics[width=\columnwidth, height=0.4\textwidth, keepaspectratio]{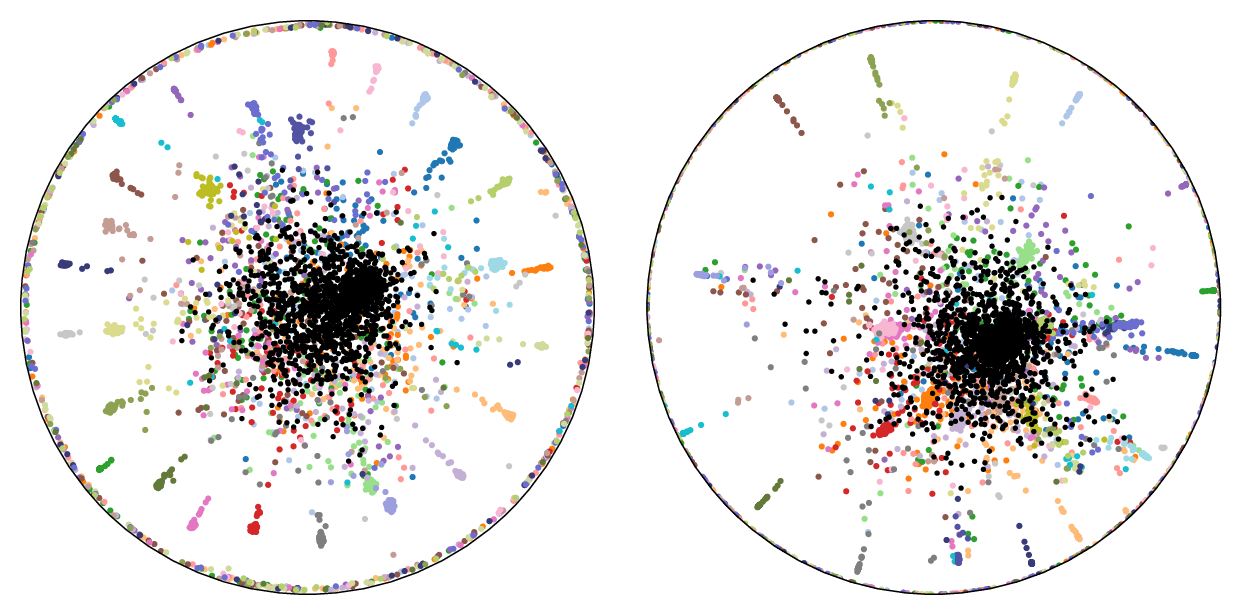}
	\caption{Continuation of Figure \ref{fig4} Hyperbolic GloVe word embeddings in the Poincar\'e disc for two additional simulations.}
	\label{figB1}
\end{figure}

\begin{figure}[t]
	\centering
	
	\parbox{0.19\textwidth}{\centering \includegraphics[width=\linewidth]{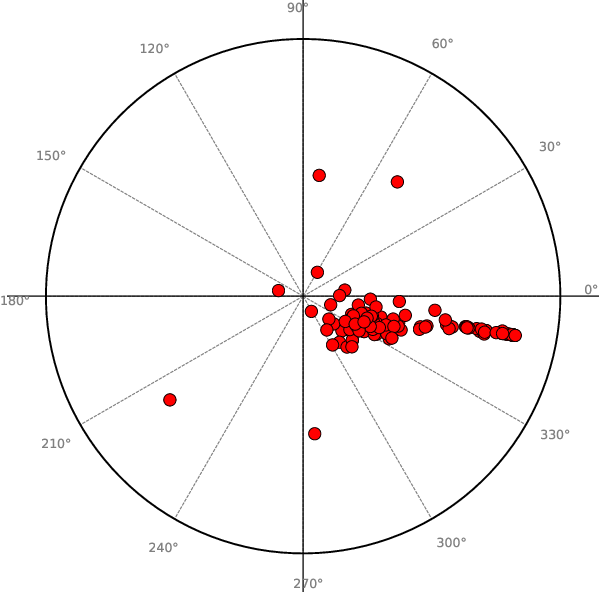}\\\scriptsize Admiration}\hfill
	\parbox{0.19\textwidth}{\centering \includegraphics[width=\linewidth]{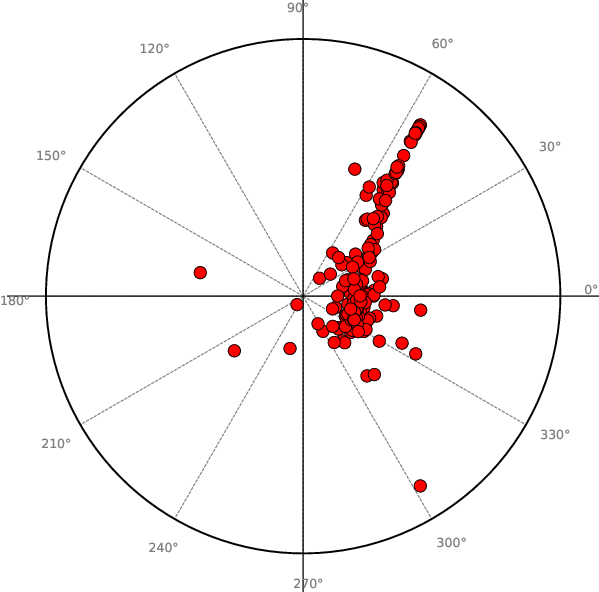}\\\scriptsize Amusement}\hfill
	\parbox{0.19\textwidth}{\centering \includegraphics[width=\linewidth]{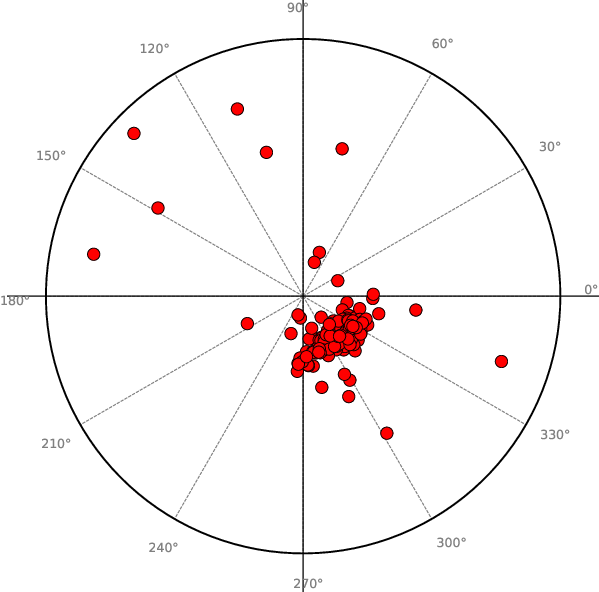}\\\scriptsize Anger}\hfill
	\parbox{0.19\textwidth}{\centering \includegraphics[width=\linewidth]{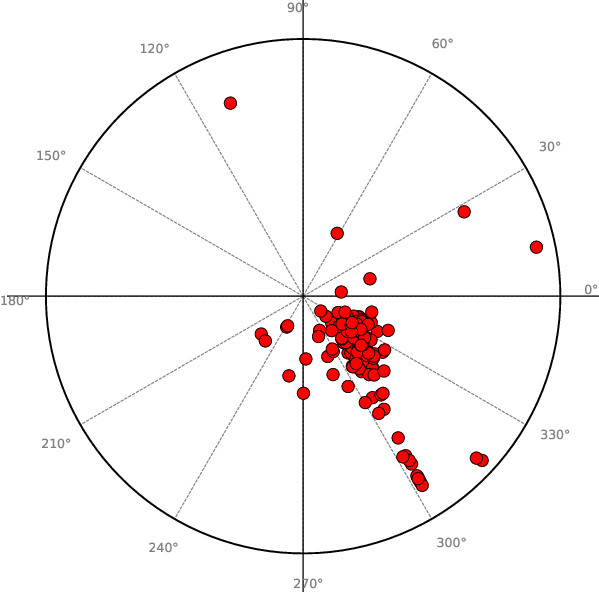}\\\scriptsize Annoyance}\hfill
	\parbox{0.19\textwidth}{\centering \includegraphics[width=\linewidth]{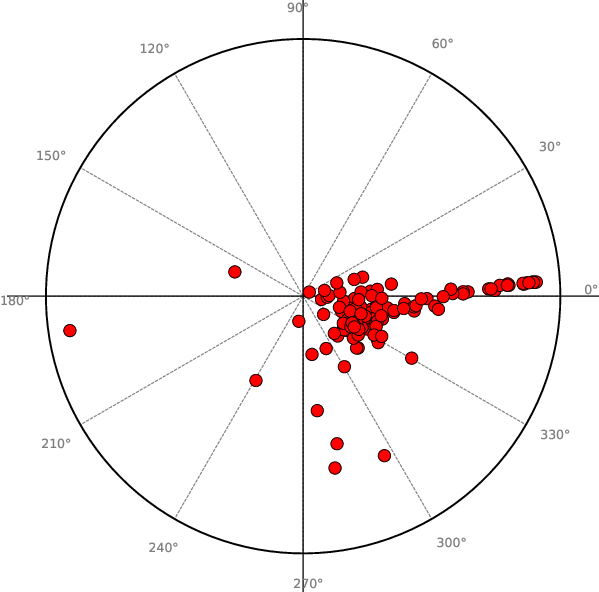}\\\scriptsize Approval}
	
	\vspace{0.3mm}
	
	\parbox{0.19\textwidth}{\centering \includegraphics[width=\linewidth]{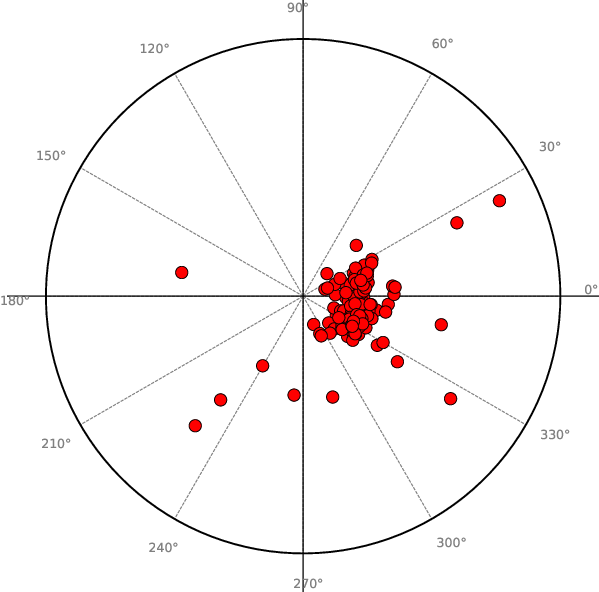}\\\scriptsize Caring}\hfill
	\parbox{0.19\textwidth}{\centering \includegraphics[width=\linewidth]{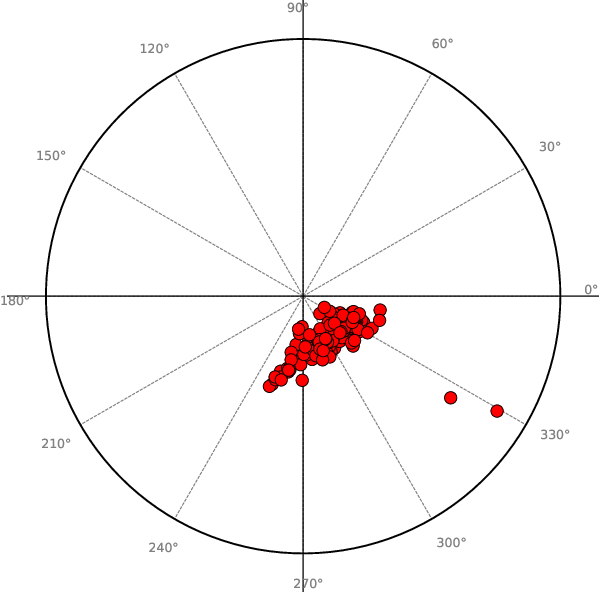}\\\scriptsize Confusion}\hfill
	\parbox{0.19\textwidth}{\centering \includegraphics[width=\linewidth]{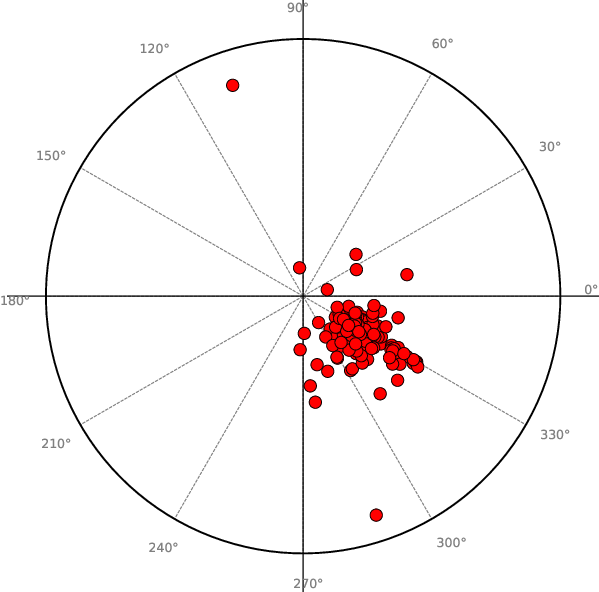}\\\scriptsize Curiosity}\hfill
	\parbox{0.19\textwidth}{\centering \includegraphics[width=\linewidth]{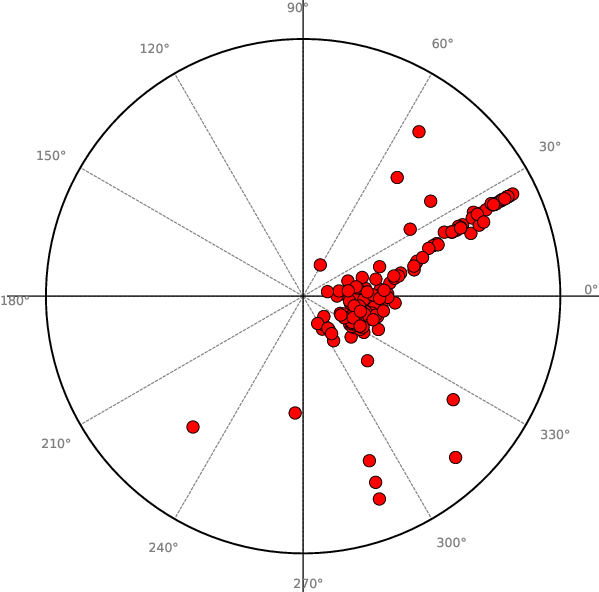}\\\scriptsize Desire}\hfill
	\parbox{0.19\textwidth}{\centering \includegraphics[width=\linewidth]{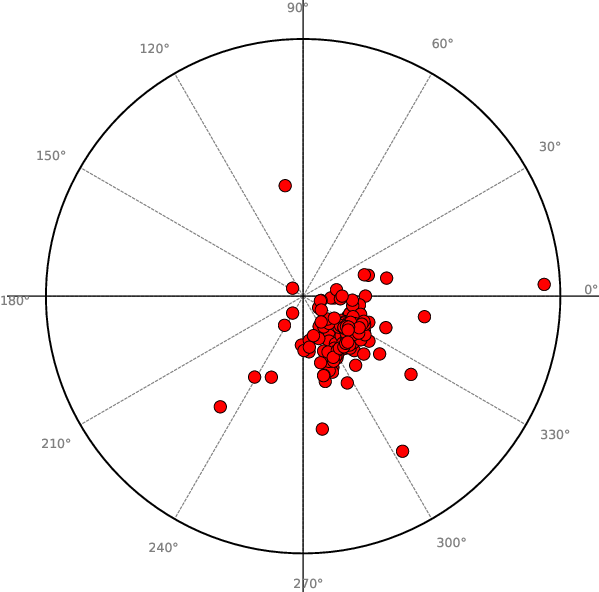}\\\scriptsize Disappointment}
	
	\vspace{0.3mm}
	
	\parbox{0.19\textwidth}{\centering \includegraphics[width=\linewidth]{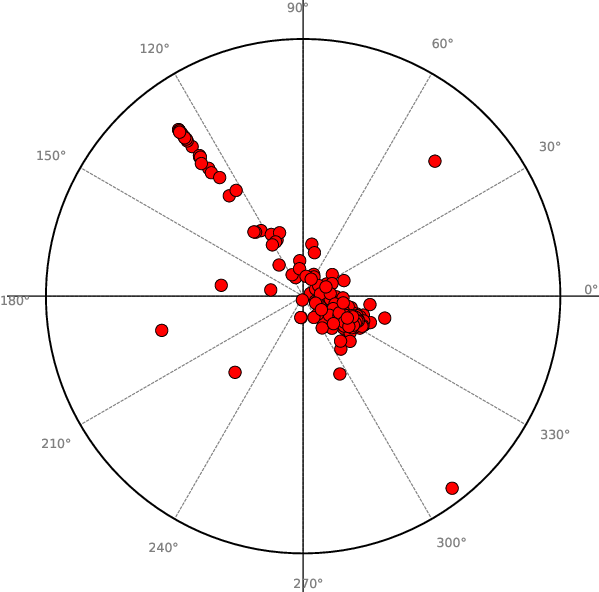}\\\scriptsize Disapproval}\hfill
	\parbox{0.19\textwidth}{\centering \includegraphics[width=\linewidth]{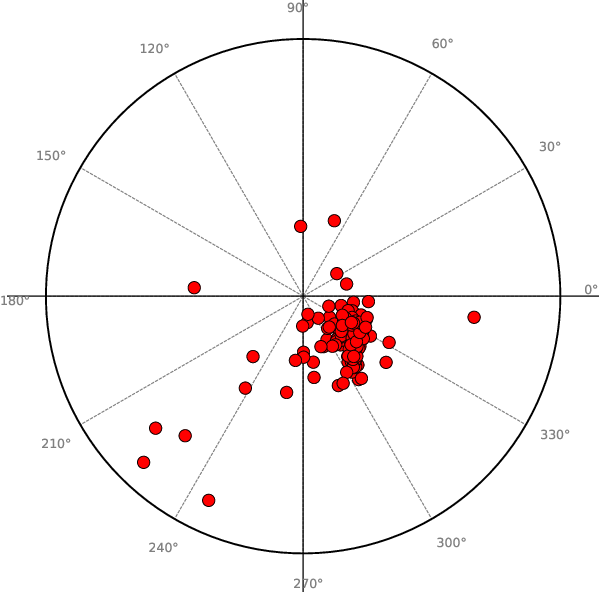}\\\scriptsize Disgust}\hfill
	\parbox{0.19\textwidth}{\centering \includegraphics[width=\linewidth]{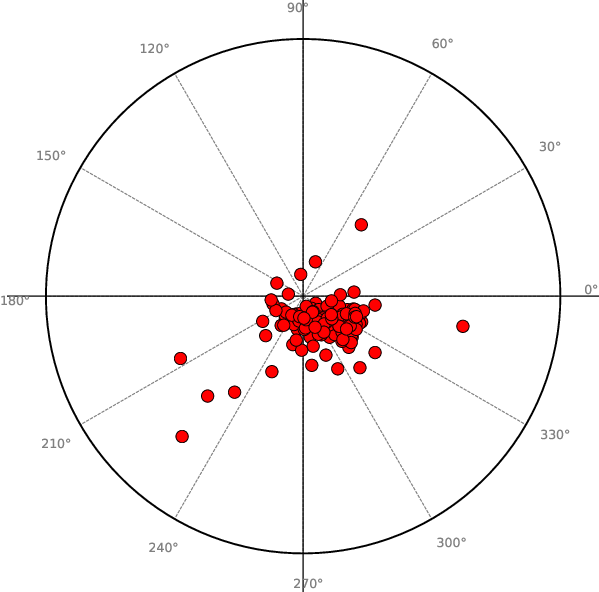}\\\scriptsize Embarrassment}\hfill
	\parbox{0.19\textwidth}{\centering \includegraphics[width=\linewidth]{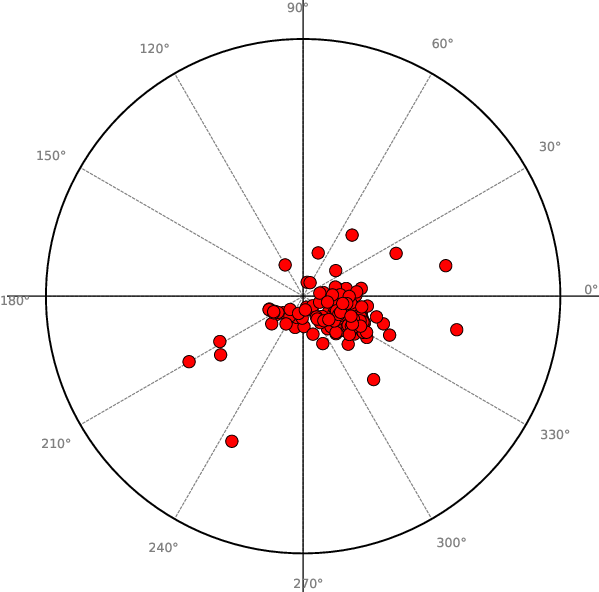}\\\scriptsize Excitement}\hfill
	\parbox{0.19\textwidth}{\centering \includegraphics[width=\linewidth]{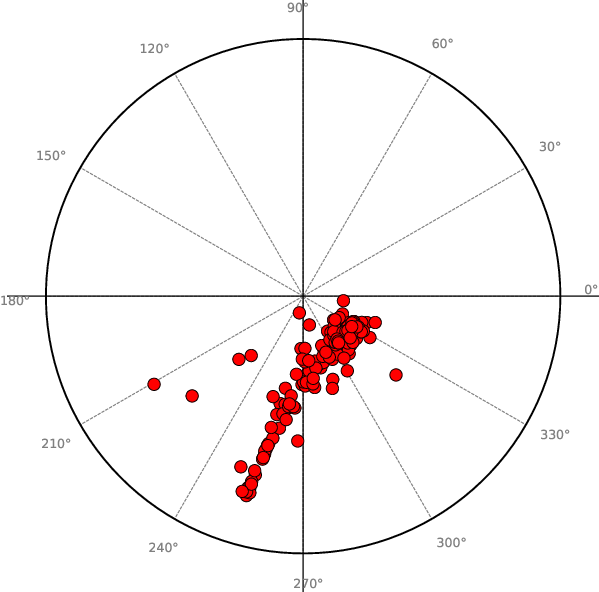}\\\scriptsize Fear}
	
	\vspace{0.3mm}
	
	\parbox{0.19\textwidth}{\centering \includegraphics[width=\linewidth]{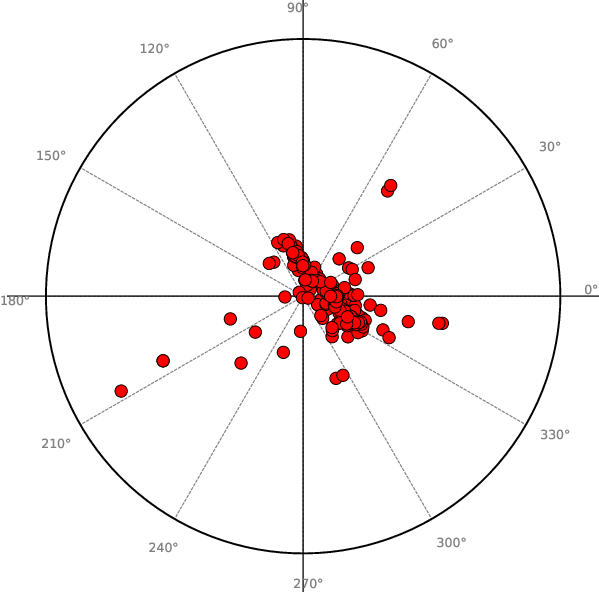}\\\scriptsize Gratitude}\hfill
	\parbox{0.19\textwidth}{\centering \includegraphics[width=\linewidth]{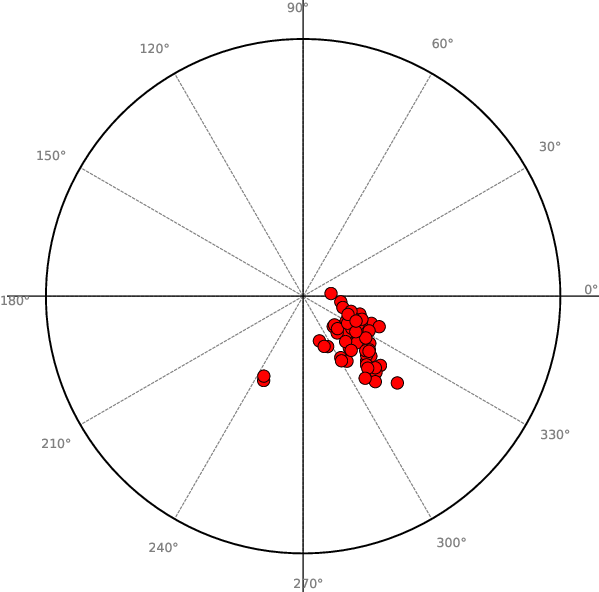}\\\scriptsize Grief}\hfill
	\parbox{0.19\textwidth}{\centering \includegraphics[width=\linewidth]{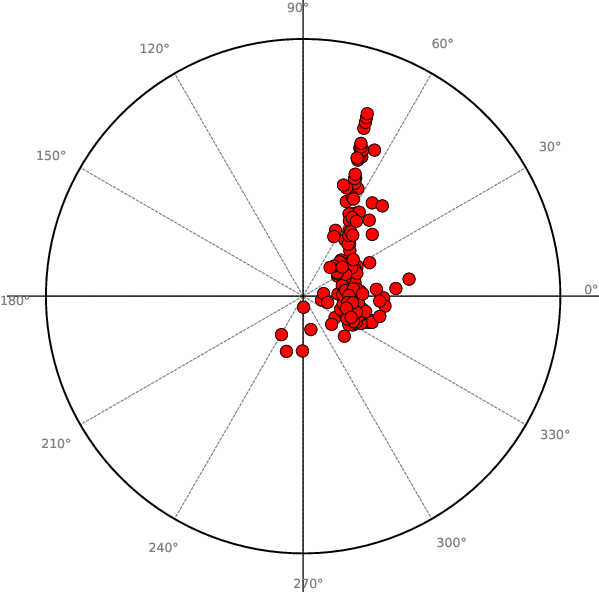}\\\scriptsize Joy}\hfill
	\parbox{0.19\textwidth}{\centering \includegraphics[width=\linewidth]{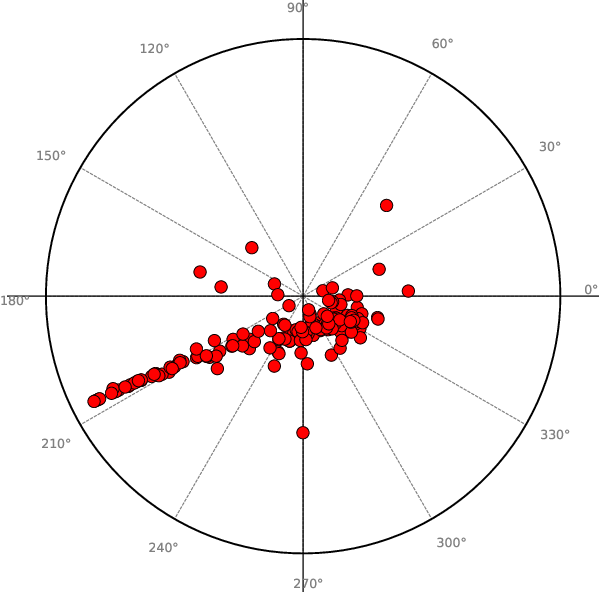}\\\scriptsize Love}\hfill
	\parbox{0.19\textwidth}{\centering \includegraphics[width=\linewidth]{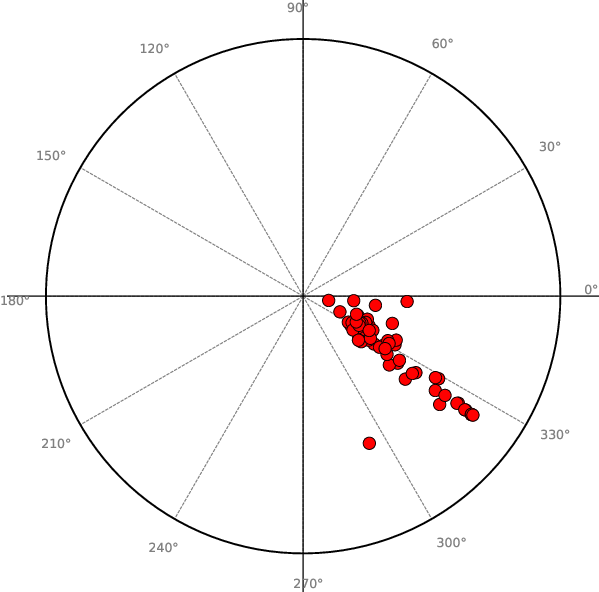}\\\scriptsize Nervousness}
	
	\vspace{0.3mm}
	
	\parbox{0.19\textwidth}{\centering \includegraphics[width=\linewidth]{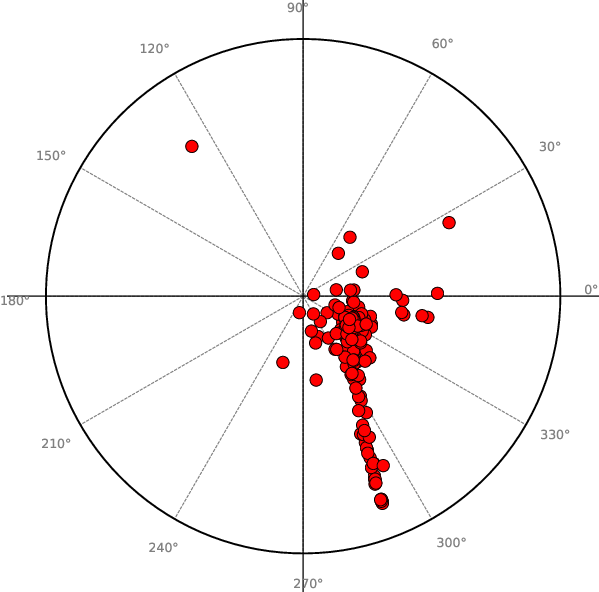}\\\scriptsize Optimism}\hfill
	\parbox{0.19\textwidth}{\centering \includegraphics[width=\linewidth]{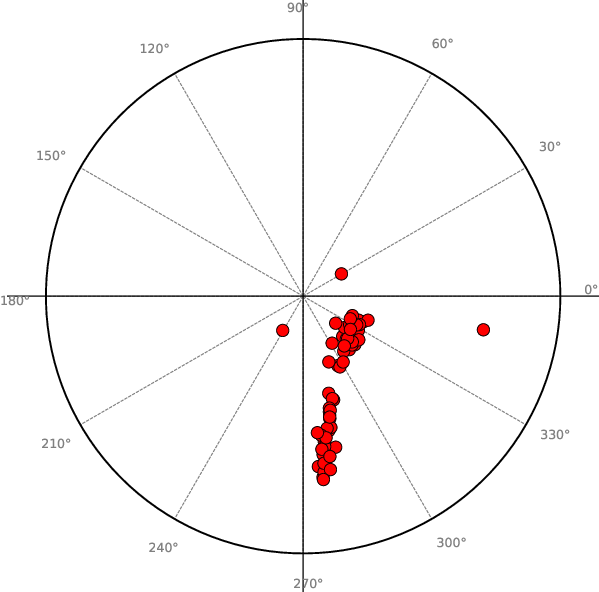}\\\scriptsize Pride}\hfill
	\parbox{0.19\textwidth}{\centering \includegraphics[width=\linewidth]{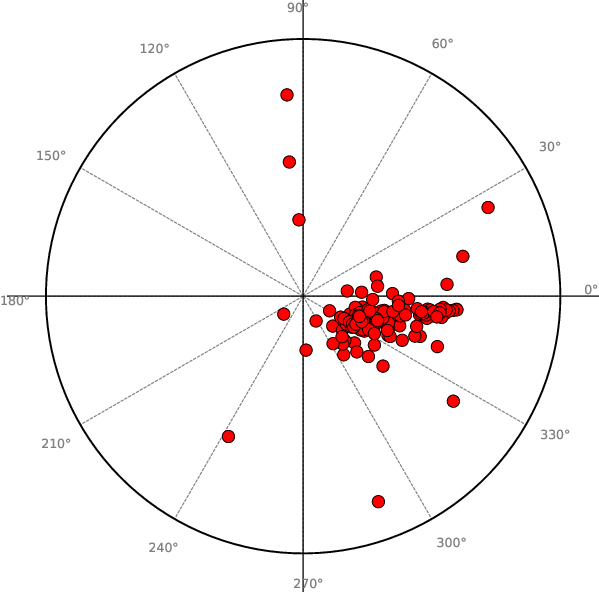}\\\scriptsize Realization}\hfill
	\parbox{0.19\textwidth}{\centering \includegraphics[width=\linewidth]{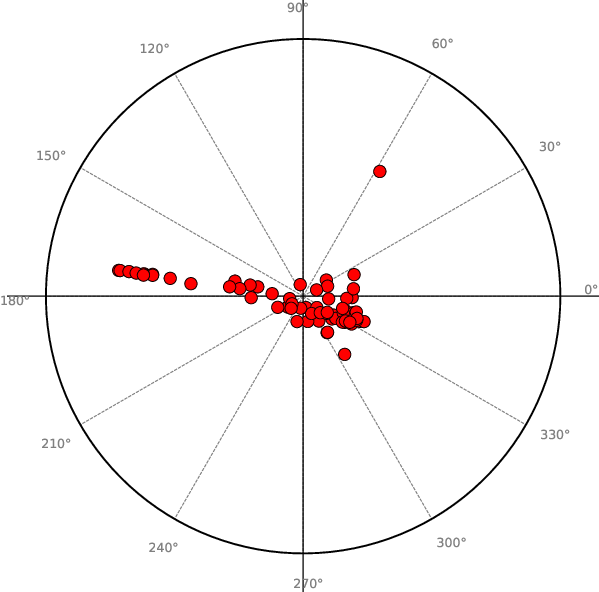}\\\scriptsize Relief}\hfill
	\parbox{0.19\textwidth}{\centering \includegraphics[width=\linewidth]{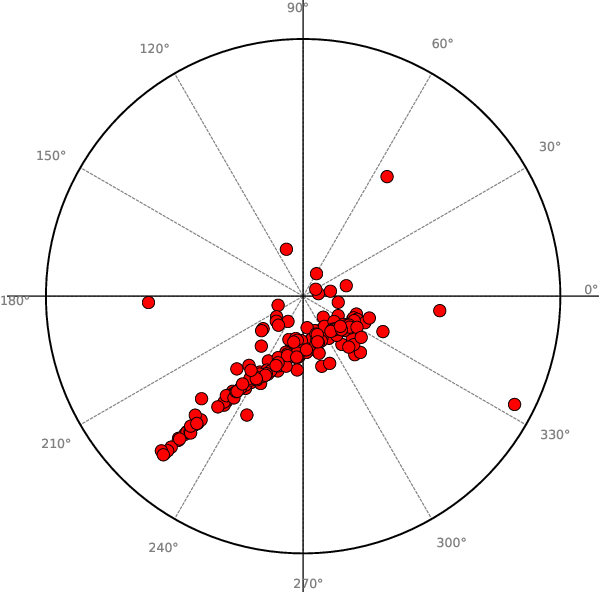}\\\scriptsize Remorse}
	
	\vspace{0.3mm}
	
	\parbox{0.19\textwidth}{}\hfill
	\parbox{0.19\textwidth}{\centering \includegraphics[width=\linewidth]{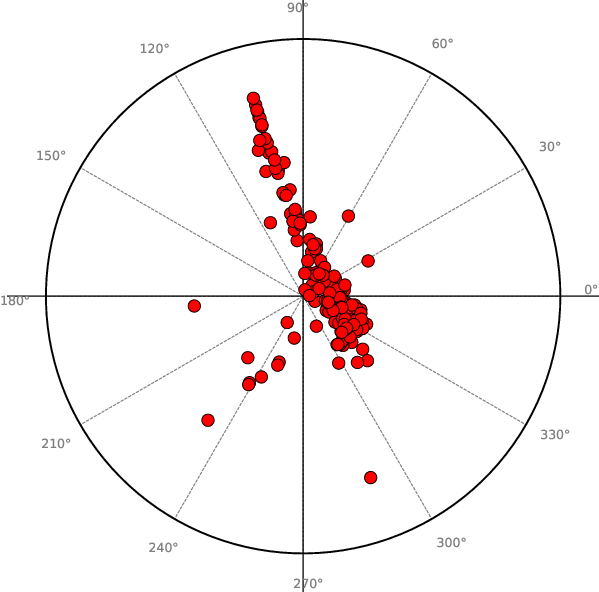}\\\scriptsize Sadness}\hfill
	\parbox{0.19\textwidth}{\centering \includegraphics[width=\linewidth]{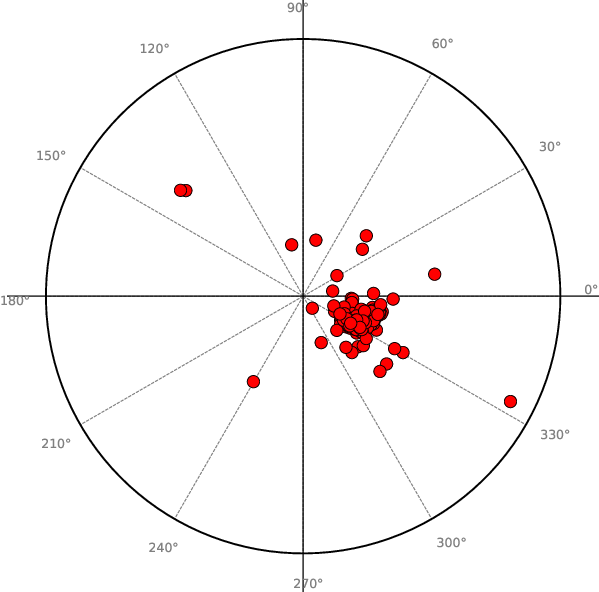}\\\scriptsize Surprise}\hfill
	\parbox{0.19\textwidth}{\centering \includegraphics[width=\linewidth]{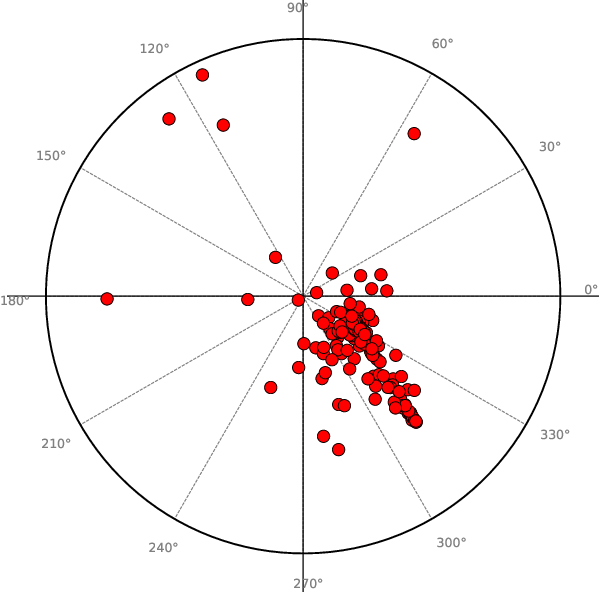}\\\scriptsize Indifference}\hfill
	\parbox{0.19\textwidth}{}
	
	\caption{Learned representations of textual messages for all 28 emotion categories. Each panel shows representations of messages annotated by the same emotion. The M\" obius correction is further applied to these representations, in order to align fingers with radii/angles.}
	\label{figB2}
\end{figure}

Here, we depict message embeddings for all of 28 emotions (see Figure \ref{figB2}). Each red dot corresponds to one message. All dots in the same disc correspond to the same emotion. The figures clearly demonstrate that the model strives to classify instances into rays (directions in the plane). Therefore, EmBolic reconsiders our assumption A2 (see subsection \ref{subsection_modeling_assumption}) on hyperbolic geometry of emotions and suggests toroidal geometry instead. It can be seen that the model creates "fingers" pointing to 28 different (almost uniformly distributed) points on the circle. Furthermore, one can observe that all fingers share the common root (epicenter) located slightly to the south-west from the center. Roughly, one fifth of instances are located along fingers. The remaining 80 \% of instances lying in the epicenter are not clearly classified by the model. This implies that the maximal accuracy that can be achieved for one embedding does not exceed 20 \%.

One might notice two slight discrepancies:

A) Fingers are not well aligned with the radii in the disc. In order to correct this, we apply the common M\" obius transformation mapping the (common) epicenter to the center of the disc. This transformation neatly aligns fingers along the radii in the disc, while preserving all distances (as it is an isometry). The corresponding modified figures are not presented here for the sake of compactness. We refer to figures in subsection \ref{subsection_text_embeddings} for three illustrative examples.

B) The fingers for a few emotions are not well pronounced. However, in other two embeddings these emotions display well pronounced finger-like structure. For each emotion we obtained at least two (out of three) well pronounced fingers.  

Finally, emphasize that repeated embeddings yield different orders of fingers. However, a strong statistical correlation between emotions becomes apparent after several embeddings.

In conclusion, the model clearly suggests classification of messages based on directions/angles. Each emotion is encoded by $k$ directions in the plane, that is - by a point on the $k$-torus. In the present study $k=3$.

\begin{figure}
	\centering
	
	\parbox{0.19\textwidth}{\centering \includegraphics[width=\linewidth]{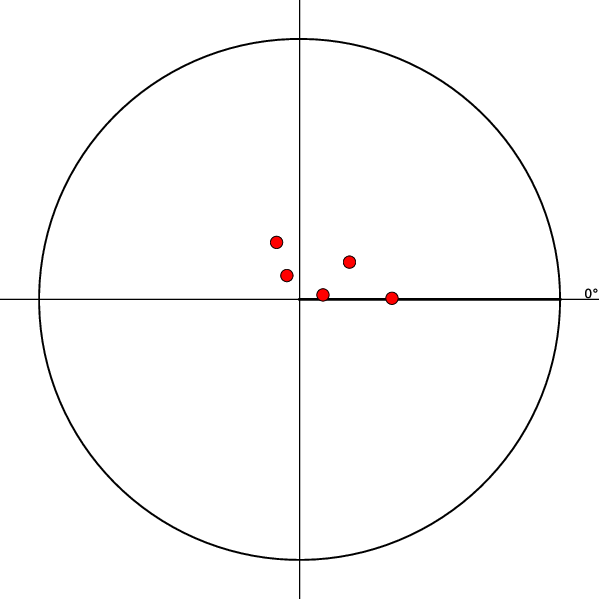}\\\scriptsize Admiration}\hfill
	\parbox{0.19\textwidth}{\centering \includegraphics[width=\linewidth]{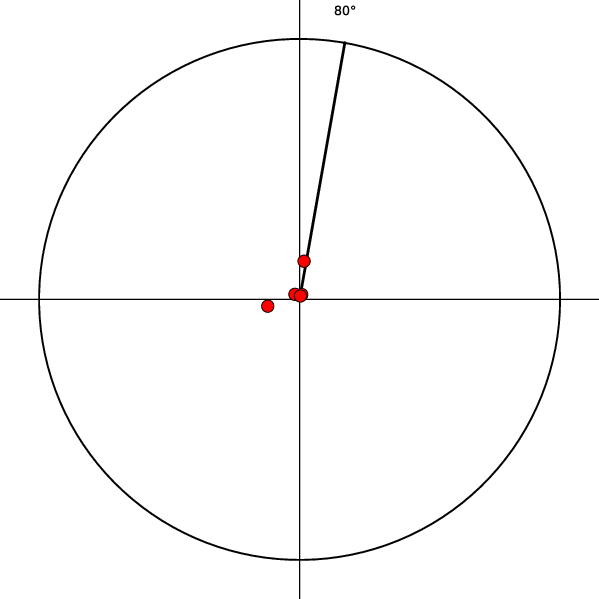}\\\scriptsize Amusement}\hfill
	\parbox{0.19\textwidth}{\centering \includegraphics[width=\linewidth]{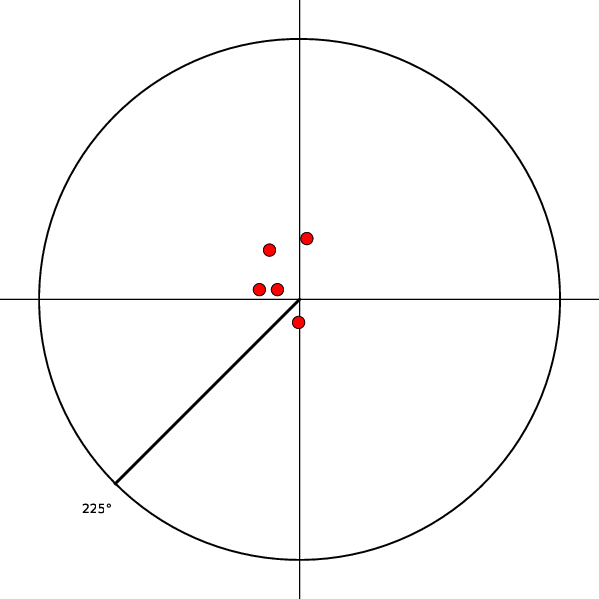}\\\scriptsize Anger}\hfill
	\parbox{0.19\textwidth}{\centering \includegraphics[width=\linewidth]{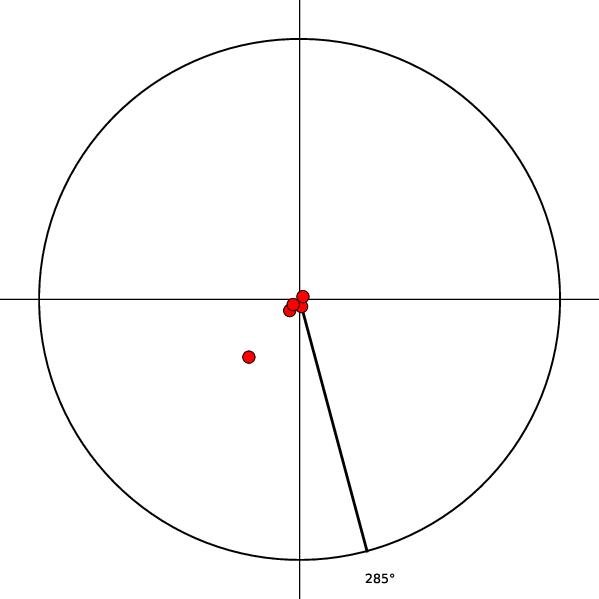}\\\scriptsize Annoyance}\hfill
	\parbox{0.19\textwidth}{\centering \includegraphics[width=\linewidth]{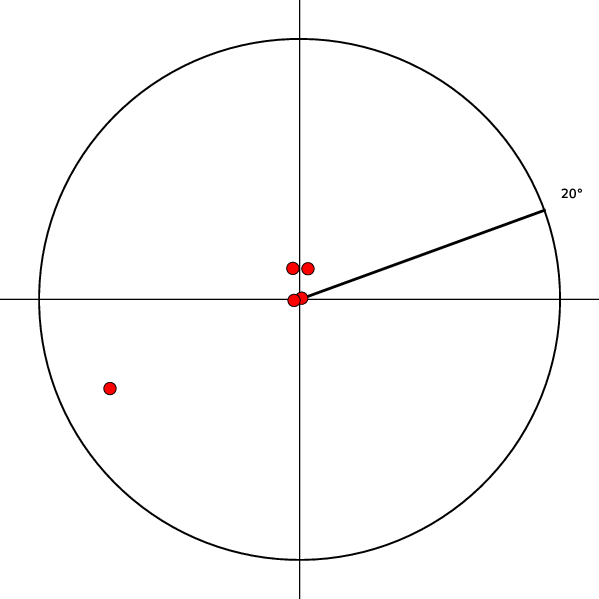}\\\scriptsize Approval}
	
	\vspace{0.3mm}
	
	\parbox{0.19\textwidth}{\centering \includegraphics[width=\linewidth]{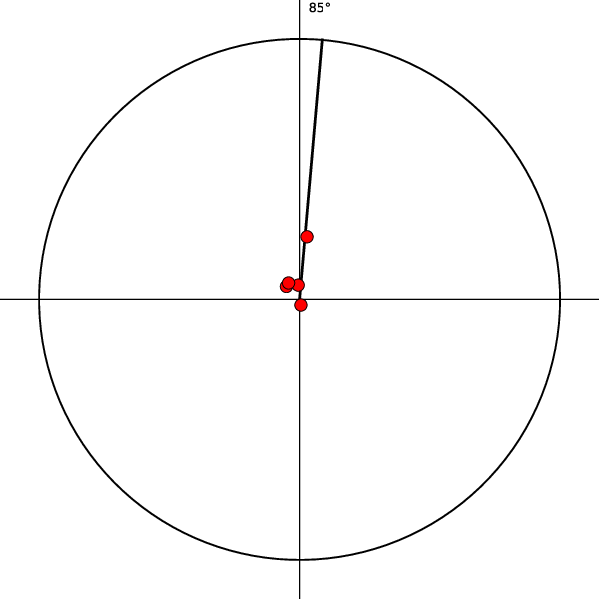}\\\scriptsize Caring}\hfill
	\parbox{0.19\textwidth}{\centering \includegraphics[width=\linewidth]{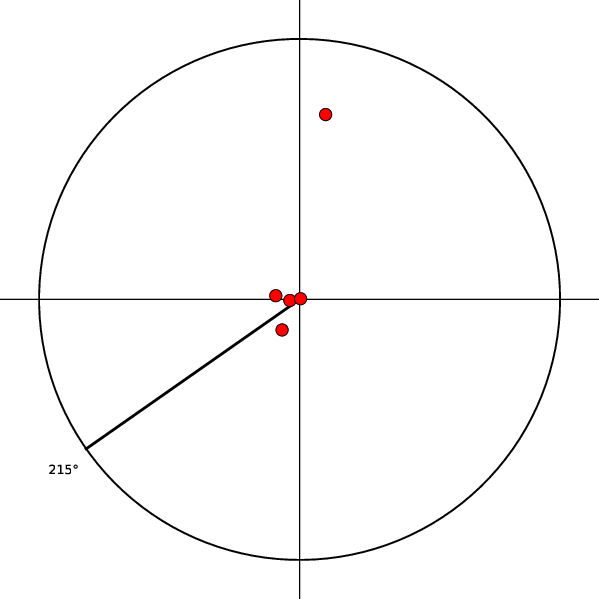}\\\scriptsize Confusion}\hfill
	\parbox{0.19\textwidth}{\centering \includegraphics[width=\linewidth]{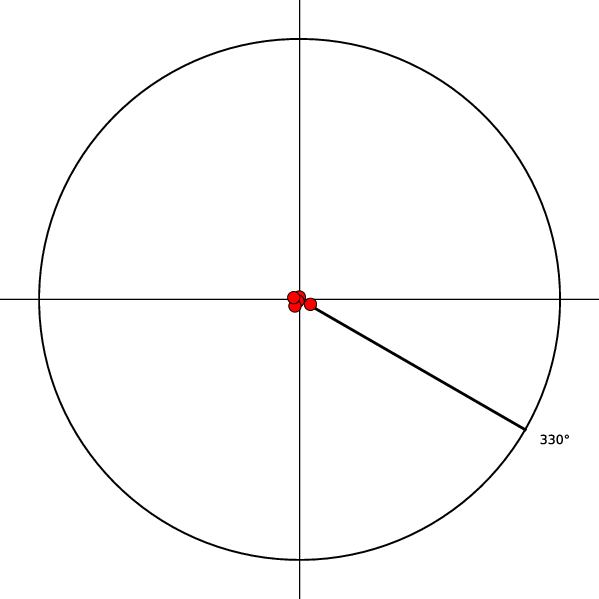}\\\scriptsize Curiosity}\hfill
	\parbox{0.19\textwidth}{\centering \includegraphics[width=\linewidth]{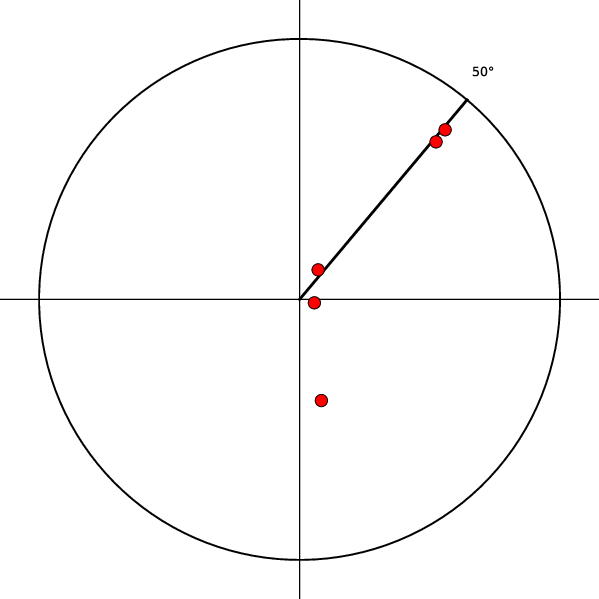}\\\scriptsize Desire}\hfill
	\parbox{0.19\textwidth}{\centering \includegraphics[width=\linewidth]{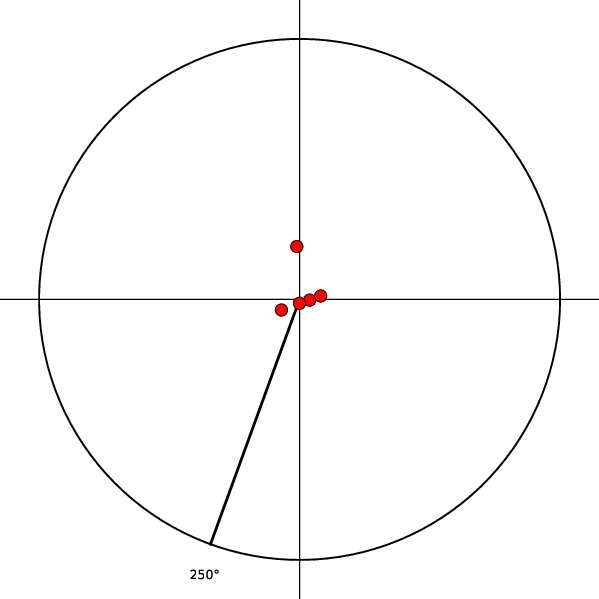}\\\scriptsize Disappointment}
	
	\vspace{0.3mm}
	
	\parbox{0.19\textwidth}{\centering \includegraphics[width=\linewidth]{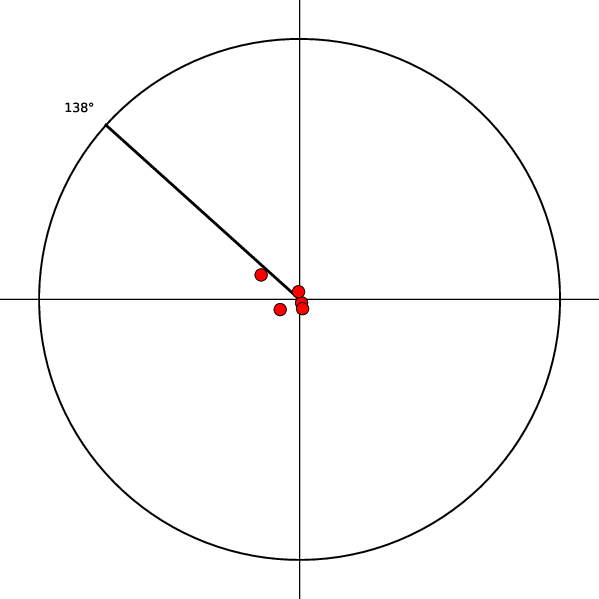}\\\scriptsize Disapproval}\hfill
	\parbox{0.19\textwidth}{\centering \includegraphics[width=\linewidth]{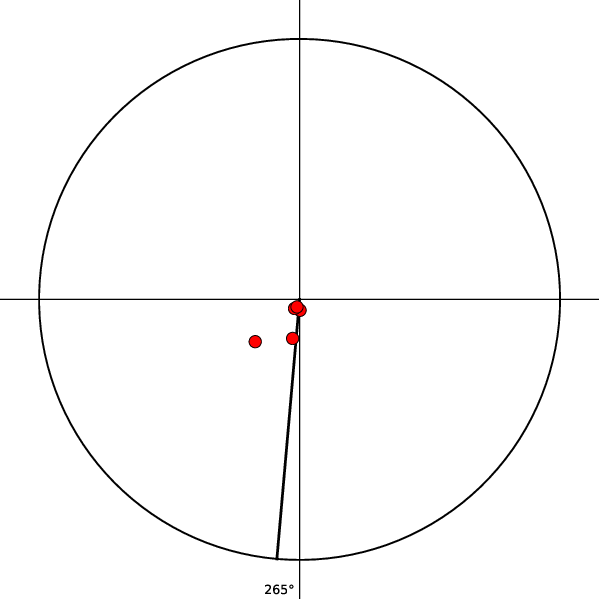}\\\scriptsize Disgust}\hfill
	\parbox{0.19\textwidth}{\centering \includegraphics[width=\linewidth]{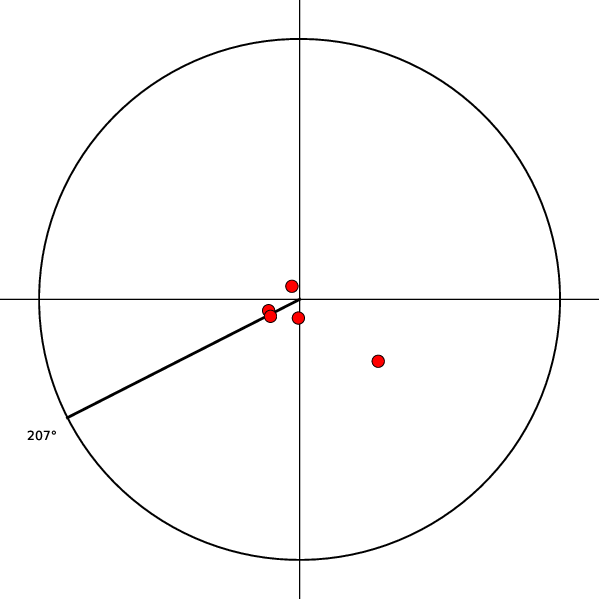}\\\scriptsize Embarrassment}\hfill
	\parbox{0.19\textwidth}{\centering \includegraphics[width=\linewidth]{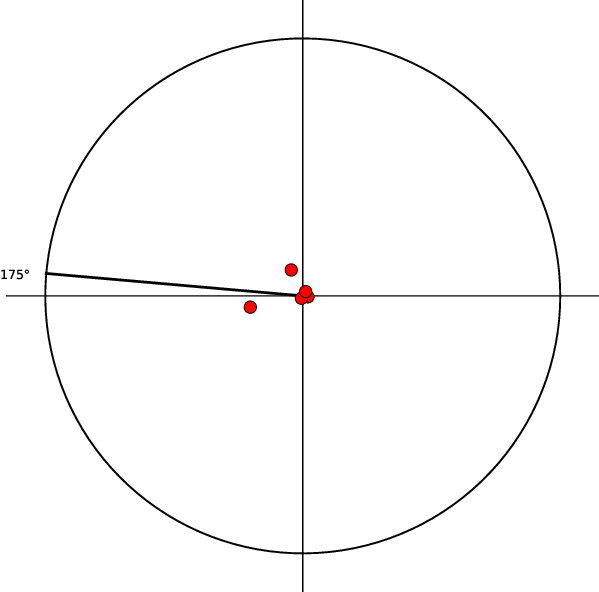}\\\scriptsize Excitement}\hfill
	\parbox{0.19\textwidth}{\centering \includegraphics[width=\linewidth]{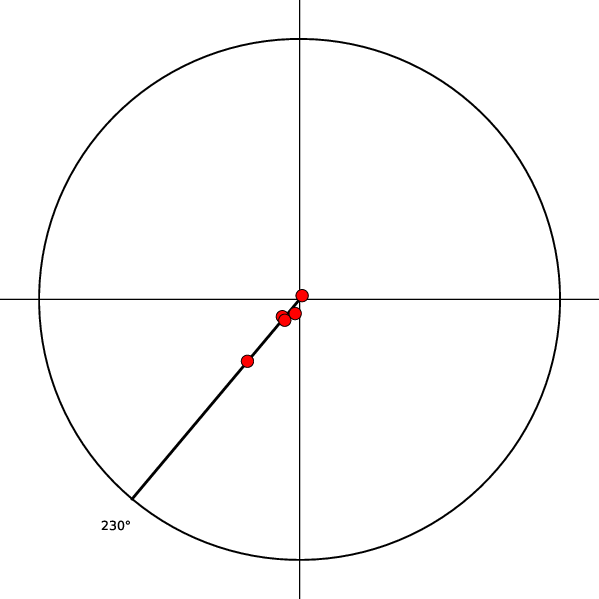}\\\scriptsize Fear}
	
	\vspace{0.3mm}
	
	\parbox{0.19\textwidth}{\centering \includegraphics[width=\linewidth]{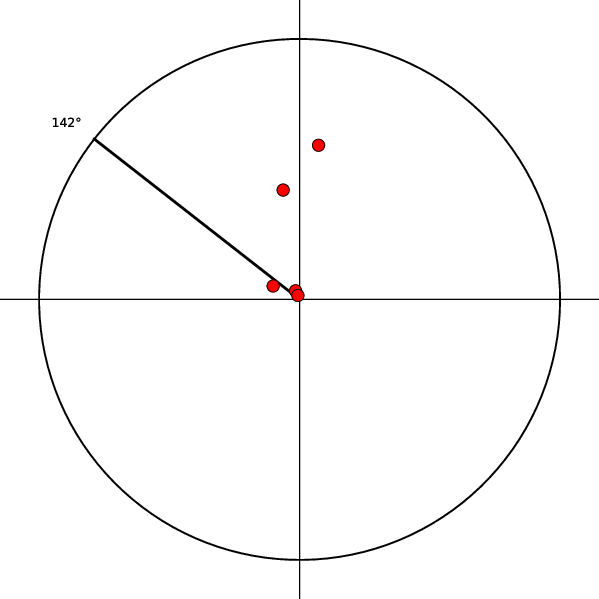}\\\scriptsize Gratitude}\hfill
	\parbox{0.19\textwidth}{\centering \includegraphics[width=\linewidth]{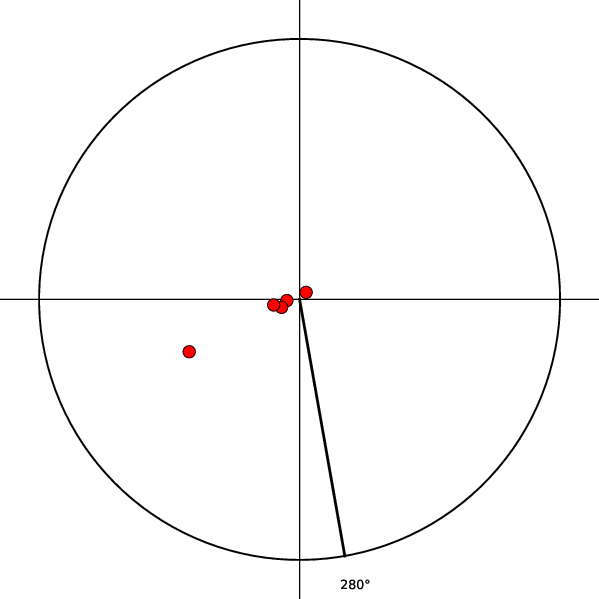}\\\scriptsize Grief}\hfill
	\parbox{0.19\textwidth}{\centering \includegraphics[width=\linewidth]{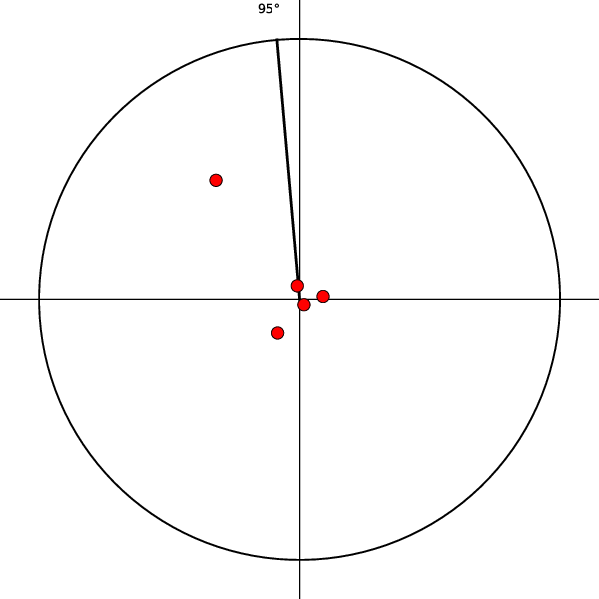}\\\scriptsize Joy}\hfill
	\parbox{0.19\textwidth}{\centering \includegraphics[width=\linewidth]{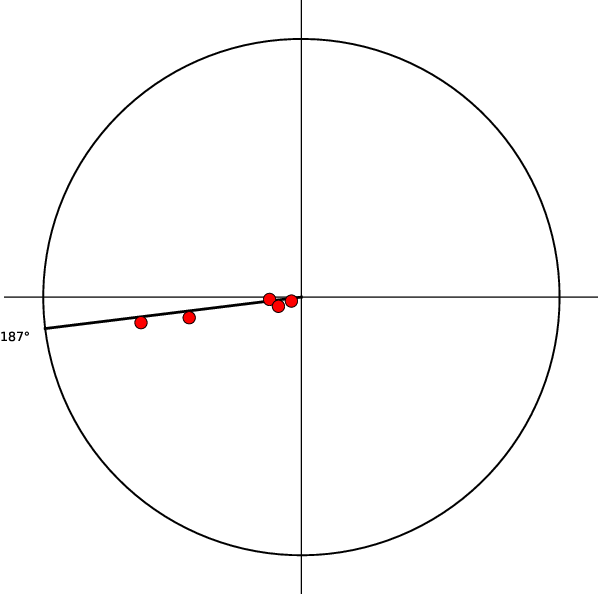}\\\scriptsize Love}\hfill
	\parbox{0.19\textwidth}{\centering \includegraphics[width=\linewidth]{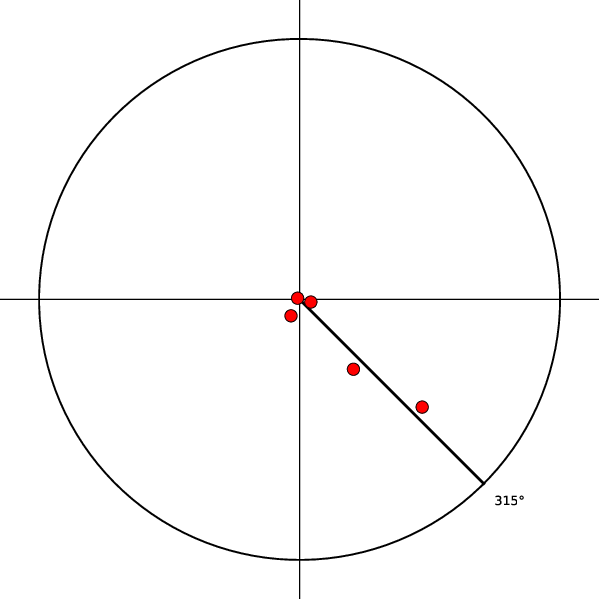}\\\scriptsize Nervousness}
	
	\vspace{0.3mm}
	
	\parbox{0.19\textwidth}{\centering \includegraphics[width=\linewidth]{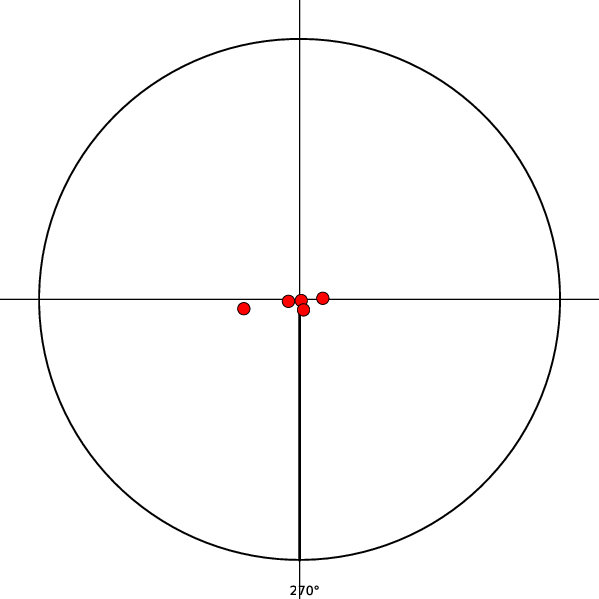}\\\scriptsize Optimism}\hfill
	\parbox{0.19\textwidth}{\centering \includegraphics[width=\linewidth]{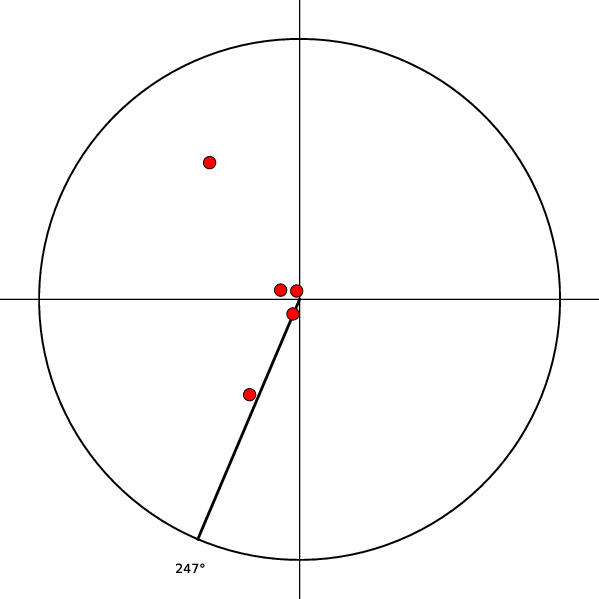}\\\scriptsize Pride}\hfill
	\parbox{0.19\textwidth}{\centering \includegraphics[width=\linewidth]{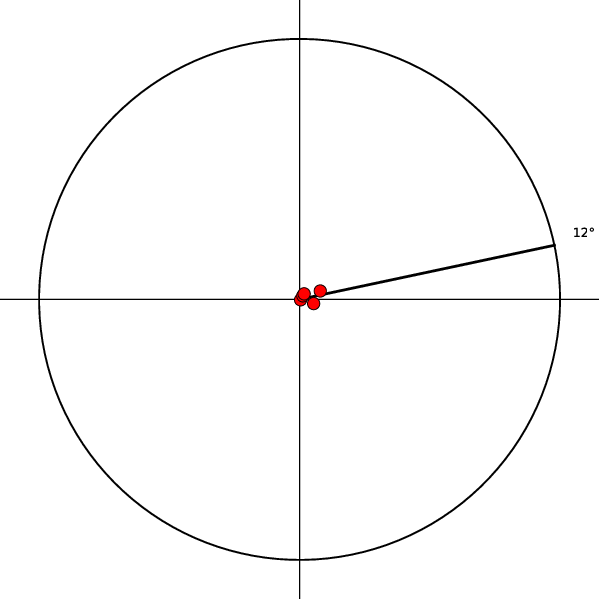}\\\scriptsize Realization}\hfill
	\parbox{0.19\textwidth}{\centering \includegraphics[width=\linewidth]{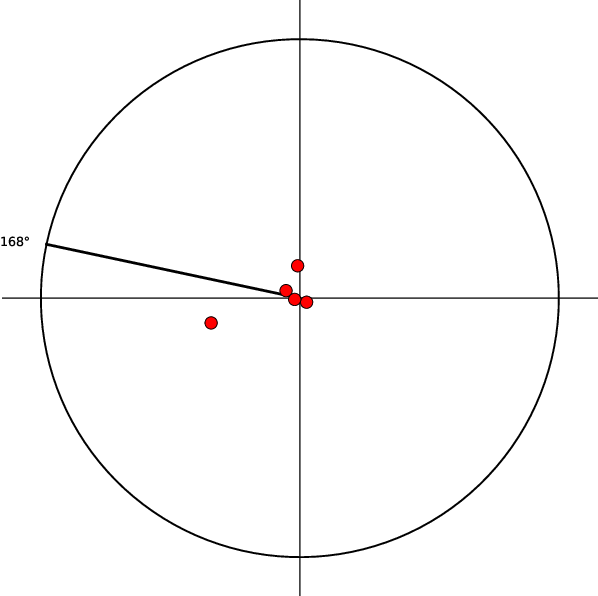}\\\scriptsize Relief}\hfill
	\parbox{0.19\textwidth}{\centering \includegraphics[width=\linewidth]{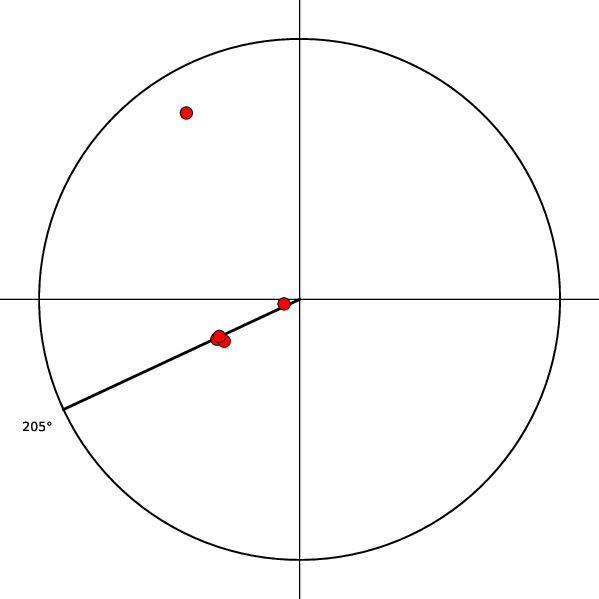}\\\scriptsize Remorse}
	
	\vspace{0.3mm}
	
	\parbox{0.19\textwidth}{}\hfill
	\parbox{0.19\textwidth}{\centering \includegraphics[width=\linewidth]{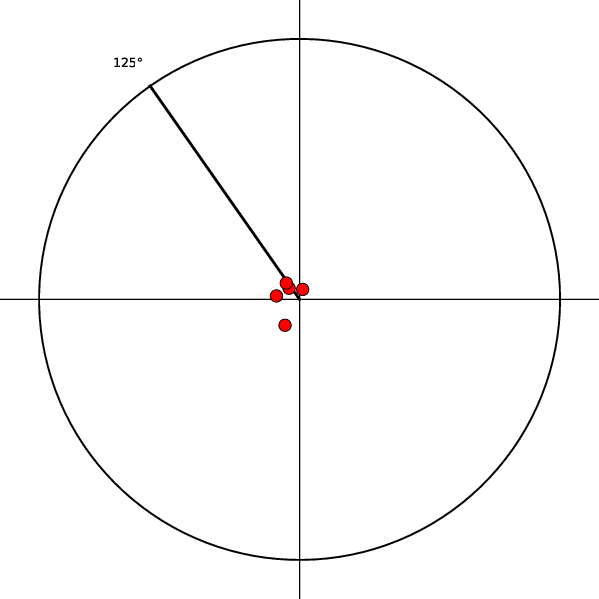}\\\scriptsize Sadness}\hfill
	\parbox{0.19\textwidth}{\centering \includegraphics[width=\linewidth]{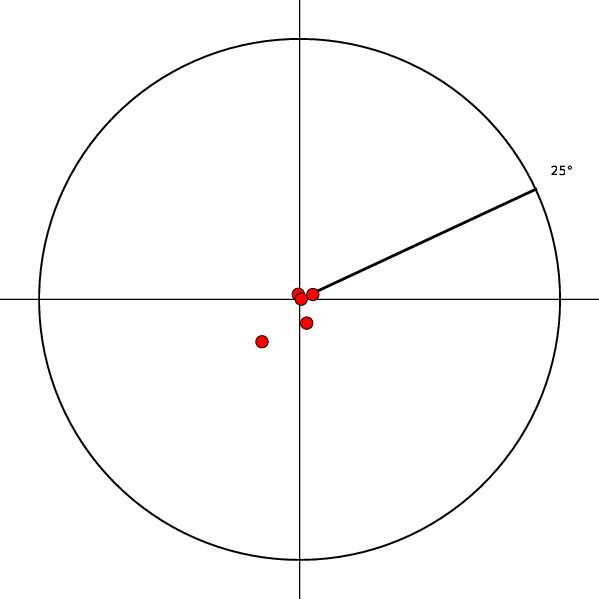}\\\scriptsize Surprise}\hfill
	\parbox{0.19\textwidth}{\centering \includegraphics[width=\linewidth]{
			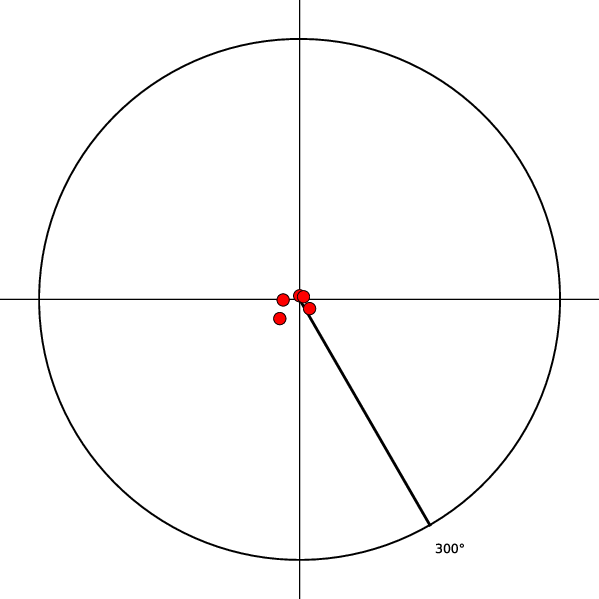}\\\scriptsize Indifference}\hfill
	\parbox{0.19\textwidth}{}
	
	\caption{Validation on the test set. Representations of test messages for 28 emotions and the radii encoding these emotions. Correct predictions correspond to the points lying on the radii. Representations shown for one embedding only. The final prediction is made based on scores obtained for three embeddings.}
	\label{figB3}
\end{figure}

Figure \ref{figB3} is a continuation of Figure \ref{fig7}. It demonstrates embeddings (5 points) of the instances from the test set together with the direction encoding the true emotion. The points lying on (or very close to) the depicted radii correspond to correctly predicted instances from the test set. Levels of confidence are encoded in distances of points to the center. 

One might notice that Figure \ref{figB3} illustrates a modest accuracy with majority of low-confident predictions and only less than 10 \% of highly confident correct predictions. However, we point out that this figure illustrates one embedding only, while accuracy consistently improves as scores from several embeddings are combined. This is substantiated in tables \ref{tab1} and \ref{tab2} in subsection \ref{validation_scores_accuracy}, where accuracy results for individual embeddings, as well as for three embeddings are provided.

\clearpage

\bibliographystyle{unsrtnat}
\bibliography{main}

\end{document}